%% file: ms.tex
\documentclass[fleqn,10pt]{wlscirep}
\usepackage[utf8]{inputenc}
\usepackage[T1]{fontenc}
\usepackage{multirow}
\usepackage{lineno}
\usepackage{booktabs}
\usepackage{tabularx}
\usepackage{subcaption}
\usepackage{todonotes}
\usepackage{color}
\usepackage{ulem}
\usepackage[official]{eurosym}
\usepackage[justification=justified,singlelinecheck=true]{caption}

\newcommand{\multilinecell}[1]{\begin{tabular}[c]{@{}c@{}}#1\end{tabular}}
\newcommand{\etal}[1]{\textit{et al.}}

\title{Critical Evaluation of Deep Neural Networks for Wrist Fracture Detection}

\author[1, +, *]{Abu Mohammed Raisuddin}
\author[1,2,+,]{Elias Vaattovaara}
\author[1,2]{Mika Nevalainen}
\author[2]{Marko Nikki}
\author[2]{Elina J{\"a}rvenp{\"a}{\"a}}
\author[2]{Kaisa Makkonen}
\author[1,2]{Pekka Pinola}
\author[1,4]{Tuula Palsio}
\author[1]{Arttu Niemensivu}
\author[1,2]{Osmo Tervonen}
\author[1,2,3]{Aleksei Tiulpin}
\affil[1]{University of Oulu, Oulu, Finland}
\affil[2]{Oulu University Hospital, Oulu, Finland}
\affil[3]{Ailean Technologies Oy, Oulu, Finland}
\affil[4]{City of Oulu, Oulu, Finland}

\affil[*]{abu.raisuddin@oulu.fi}
\affil[+]{these authors contributed equally to this work}

\begin{abstract}
\noindent
Wrist Fracture is the most common type of fracture with a high incidence rate. Conventional radiography (i.e. X-ray imaging) is used for wrist fracture detection routinely, but occasionally fracture delineation poses issues and an additional confirmation by computed tomography (CT) is needed for diagnosis. Recent advances in the field of Deep Learning (DL), a subfield of Artificial Intelligence (AI), have shown that wrist fracture detection can be automated using Convolutional Neural Networks. However, previous studies did not pay close attention to the difficult cases which can only be confirmed via CT imaging. In this study, we have developed and analyzed a state-of-the-art DL-based pipeline for wrist (distal radius) fracture detection -- DeepWrist, and evaluated it against one general population test set, and one challenging test set comprising only cases requiring confirmation by CT. Our results reveal that a typical state-of-the-art approach, such as DeepWrist, while having a near-perfect performance on the general independent test set, has a substantially lower performance on the challenging test set -- average precision of 0.99 (0.99-0.99) vs 0.64 (0.46-0.83), respectively. Similarly, the area under the ROC curve was of 0.99 (0.98-0.99) vs 0.84 (0.72-0.93), respectively. Our findings highlight the importance of a meticulous analysis of DL-based models before clinical use, and unearth the need for more challenging settings for testing medical AI systems.

\end{abstract}
\begin{document}

\flushbottom
\maketitle

\thispagestyle{empty}

\input{section/introduction}
\input{section/methods}

\input{section/result}
\input{section/discussion}

\section*{Acknowledgements}
This project was supported by the internal funds of the Research Unit of Medical Imaging, Physics and Technology, University of Oulu.

\section*{Additional Information}
\subsection*{Data Availability}
A Python implementation of DeepWrist is available at \url{https://github.com/MIPT-Oulu/DeepWrist}. The training and test data are not public. The repository contains Singularity and Docker containers for testing wrist radiographs in DICOM format. 

\subsection*{Author contributions statement}
A.T., E.V., O.T. and M.N designed the experiments and organized the train data collection. E.V. collected the general population test datasets, organized all test sets' annotation and provided the clinical interpretation of the findings, and participate in the initial draft of the manuscript. A.N. collected the challenging test set. M.N(2)., E.J., K.M., P.P., and T.P. annotated the test data. A.M.R. annotated anatomical landmarks, conducted the experiments, gathered and formally analyzed the results, and wrote the first draft of the manuscript. A.T. supervised the project. All authors reviewed the manuscript and participated in its preparation.

\subsection*{Competing interests}
Dr. Aleksei Tiulpin is a co-founder and a shareholder of Ailean Technologies Oy. Other authors declare no competing interests.
\bibliography{ms}

\end{document}


\normalem
\flushbottom
\maketitle
\vspace{-2cm}
\section{Dataset statistics}
\input{tables/data_stat}

\begin{figure}[ht!]
    \centering
    \includegraphics[width=1.0\linewidth]{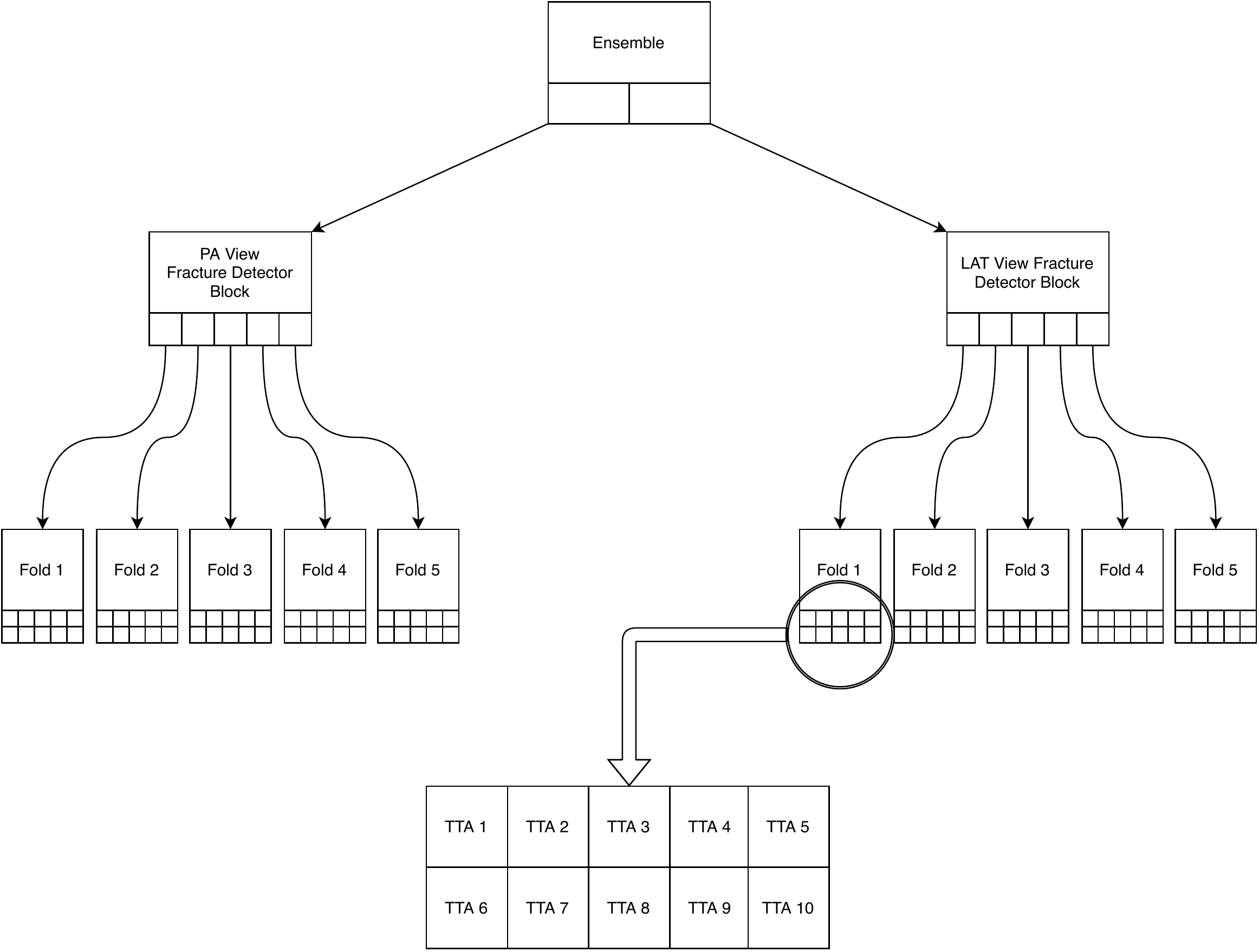}
    \caption{Structure of Both view ensemble of Fracture Detector Block. TTA stands for Test-Time Augmentation, PA for Posterioanterior, LAT for Lateral }
    \label{fig:ensemble_fd}
\end{figure}

\section{Landmark Localization}
\begin{figure}[ht!]
    \centering
    \includegraphics[width=0.5\linewidth]{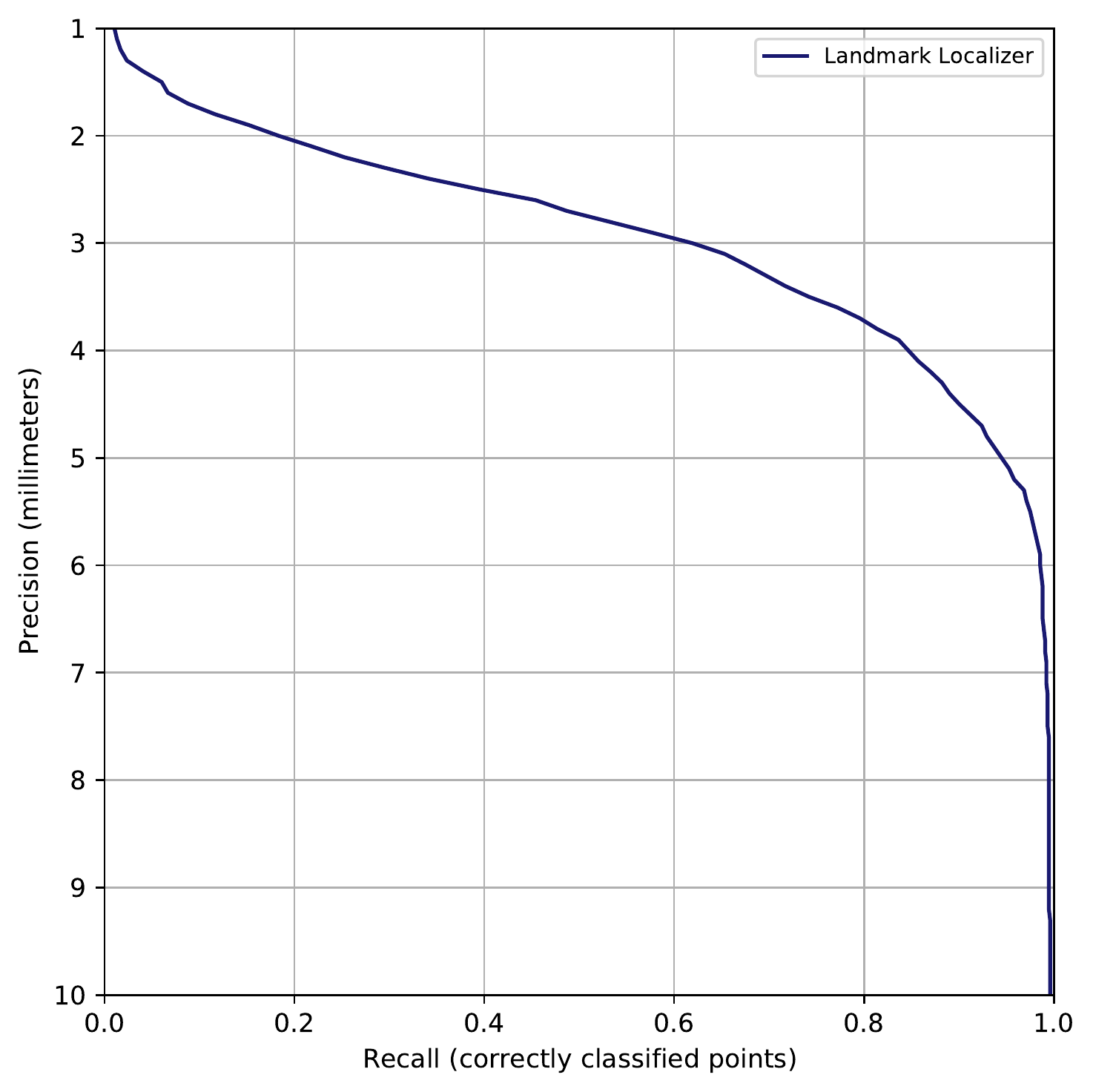}
    \caption{Precision-Recall Curve for Landmark Localizer}
    \label{fig:suppli_pr_lm}
\end{figure}

Annotations for the landmark localization were annotated by the first author. For each radiographs three anatomical landmarks were annotated. These landmarks are: top of distal ulna, top of distal radius and assumed center of the wrist for PA view and two distinguishable landmarks on top part of distal radio-ulna and the assumed center of wrist for LAT view. Since these landmarks are not exact points we did intra-rater repeatability analysis. To that end, we randomly chose 100 radiographs from fracture and normal category for both PA and LAT view totaling 400 radiographs. Then we re-annotated them without assessing how they are annotated in the first annotation. Since it is not classification, we cannot compute the Cohen's Quadratic Kappa, instead, we calculated the recall at certain precision. With respect to first annotations, the second annotations scored recall of $0.16\, (0.12 - 0.19)$ at $2mm$ precision, $0.55\, (0.50 - 0.60)$ at $4mm$ precision, $0.70\, (0.65 - 0.74)$ at $5mm$ precision. If we calculate recall for X-coordinates only, we got a recall of $0.98 \, (0.97-0.99)$ at $5mm$ precision and for Y-coordinates we got a recall of $0.87 \,(0.84 - 0.90)$ at $5mm$ precision.We visualize the Precision-Recall curve for the landmark localizer in~\autoref{fig:suppli_pr_lm}. 

\section{Inter-Rater Agreement Analysis}
\begin{figure}[ht!]
\centering

\subfloat[ \label{fig:ts1_gt_kappa}]{\includegraphics[width=.45\linewidth]{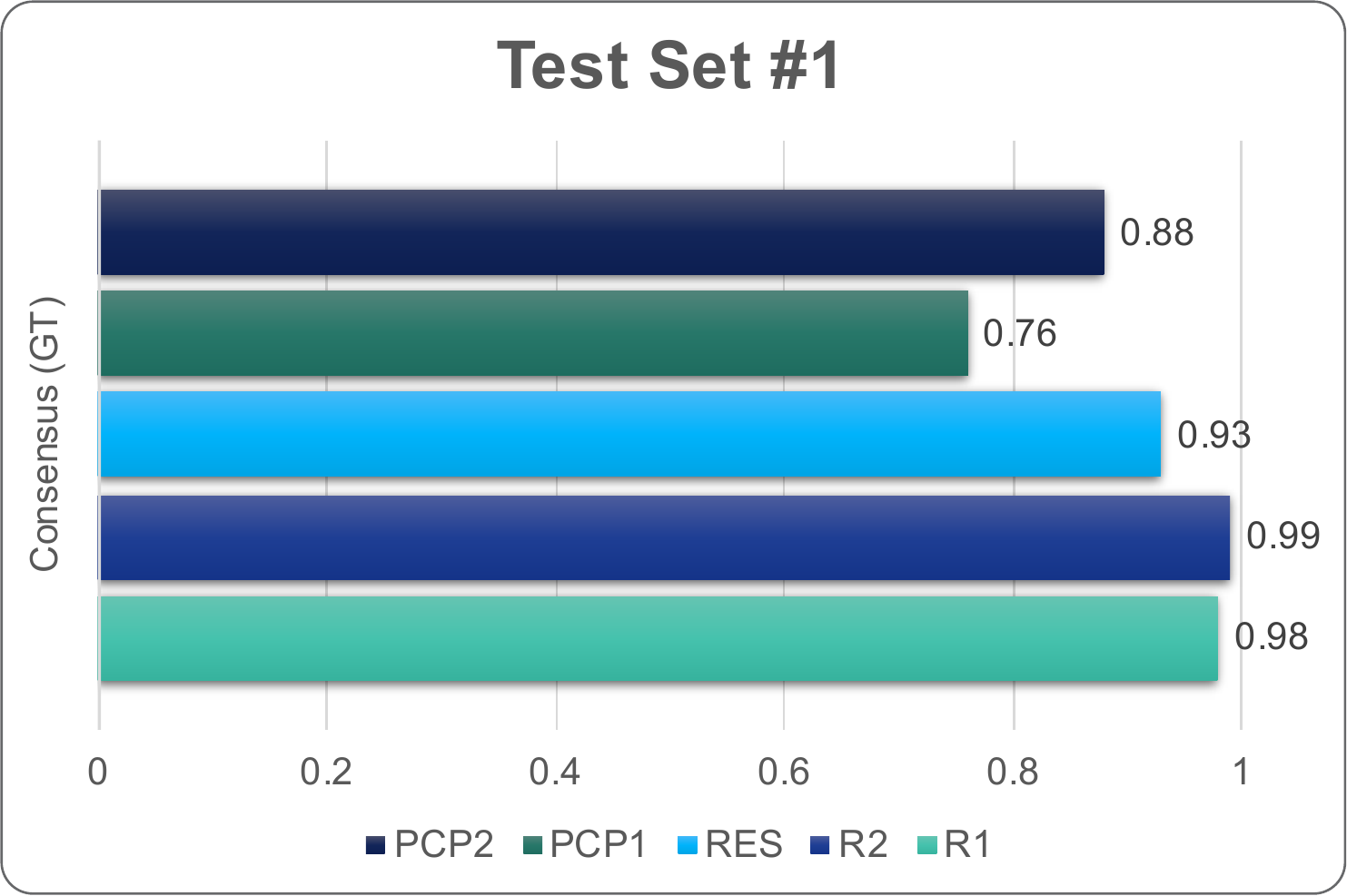}}
\hfil
\subfloat[ \label{fig:ts2_gt_kappa}]{\includegraphics[width=.45\linewidth]{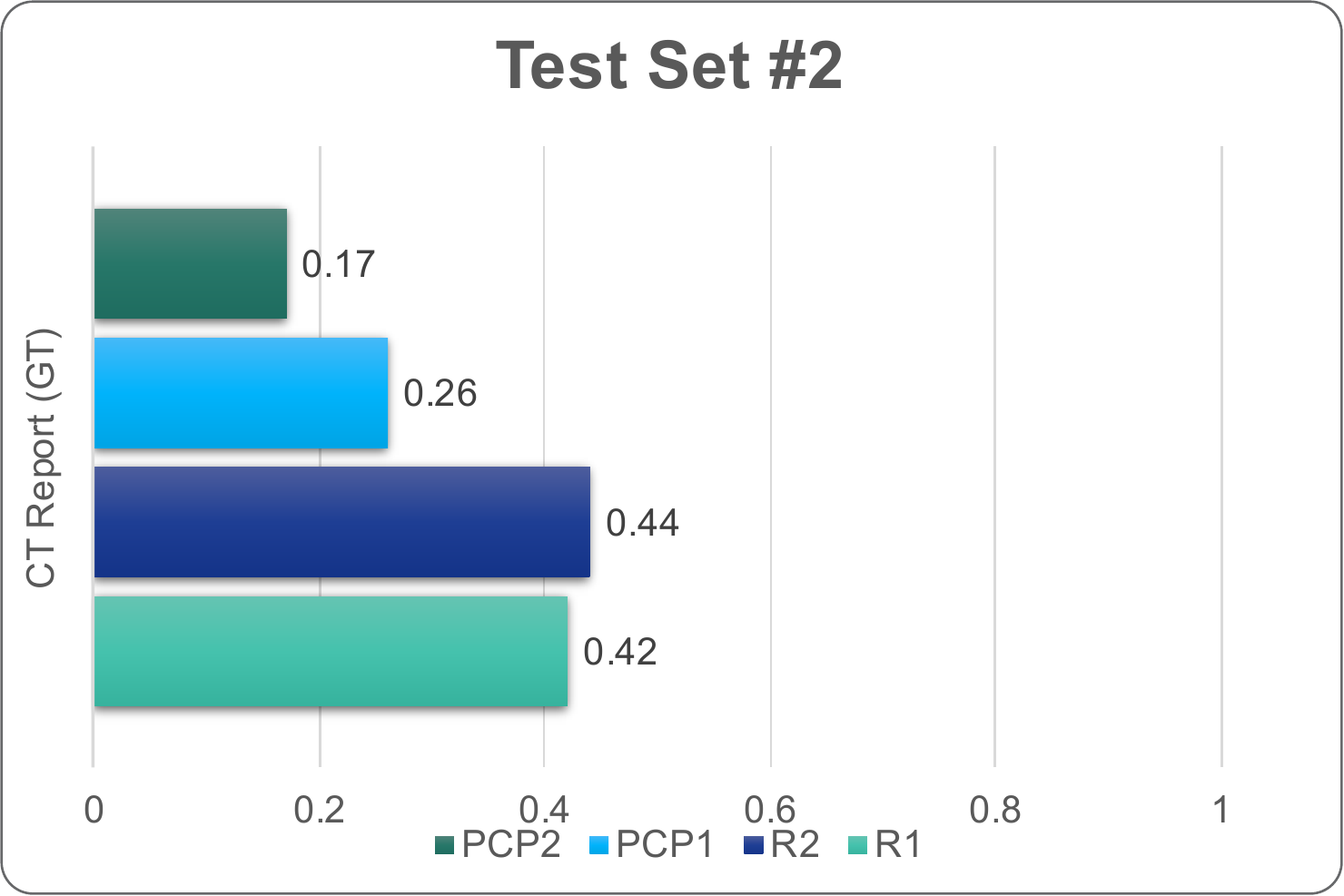}}

\subfloat[ \label{fig:ts1_pcp_kappa}]{\includegraphics[width=.45\linewidth]{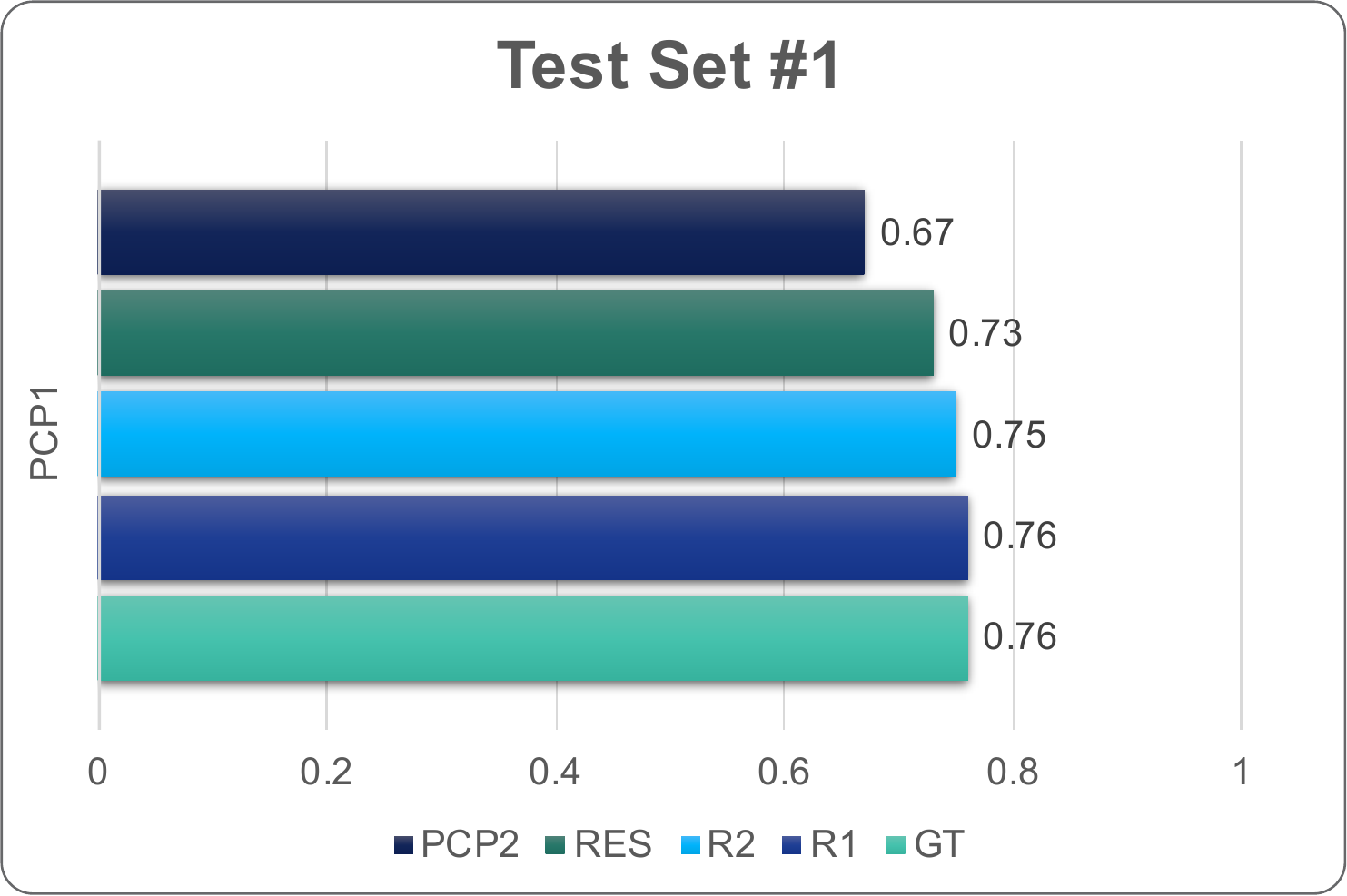}}
\hfil
\subfloat[ \label{fig:ts2_pcp_kappa}]{\includegraphics[width=.45\linewidth]{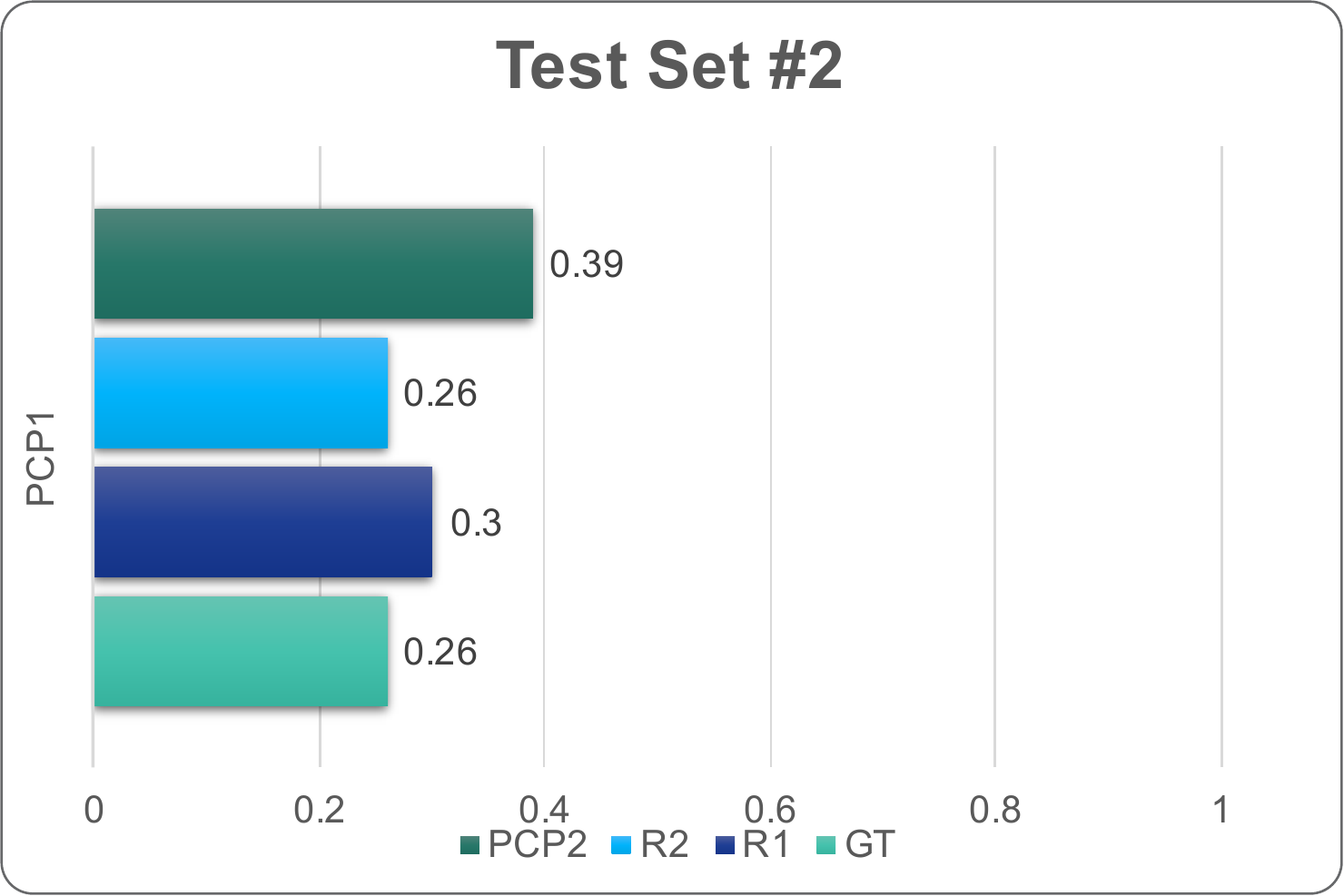}}

\caption{Inter-rater analysis using Cohen's Quadratic Kappa. (a) Agreement of Radiologist 1 (R1), Radiologist 2 (R2), Radiology Resident (RES), Primary Care Physician 1 (PCP1) and Primary Care Physician 2 (PCP2) with respect to the Ground Truth (GT) derived from consensus of two radiologists for the Test Set \#1 (b) Agreement of R1, R2, PCP1 and PCP2 with respect to GT derived from CT report for Test Set \#2. (c) Agreement of R1, R2, RES, PCP2 and GT with respect to PCP1 for Test Set \#1. (d) Agreement of R1, R2, PCP2 and GT with respect to PCP1 for Test Set \#2.}
\label{fig:kappa}
\end{figure}

\begin{figure}[ht!]
    \centering
    \includegraphics{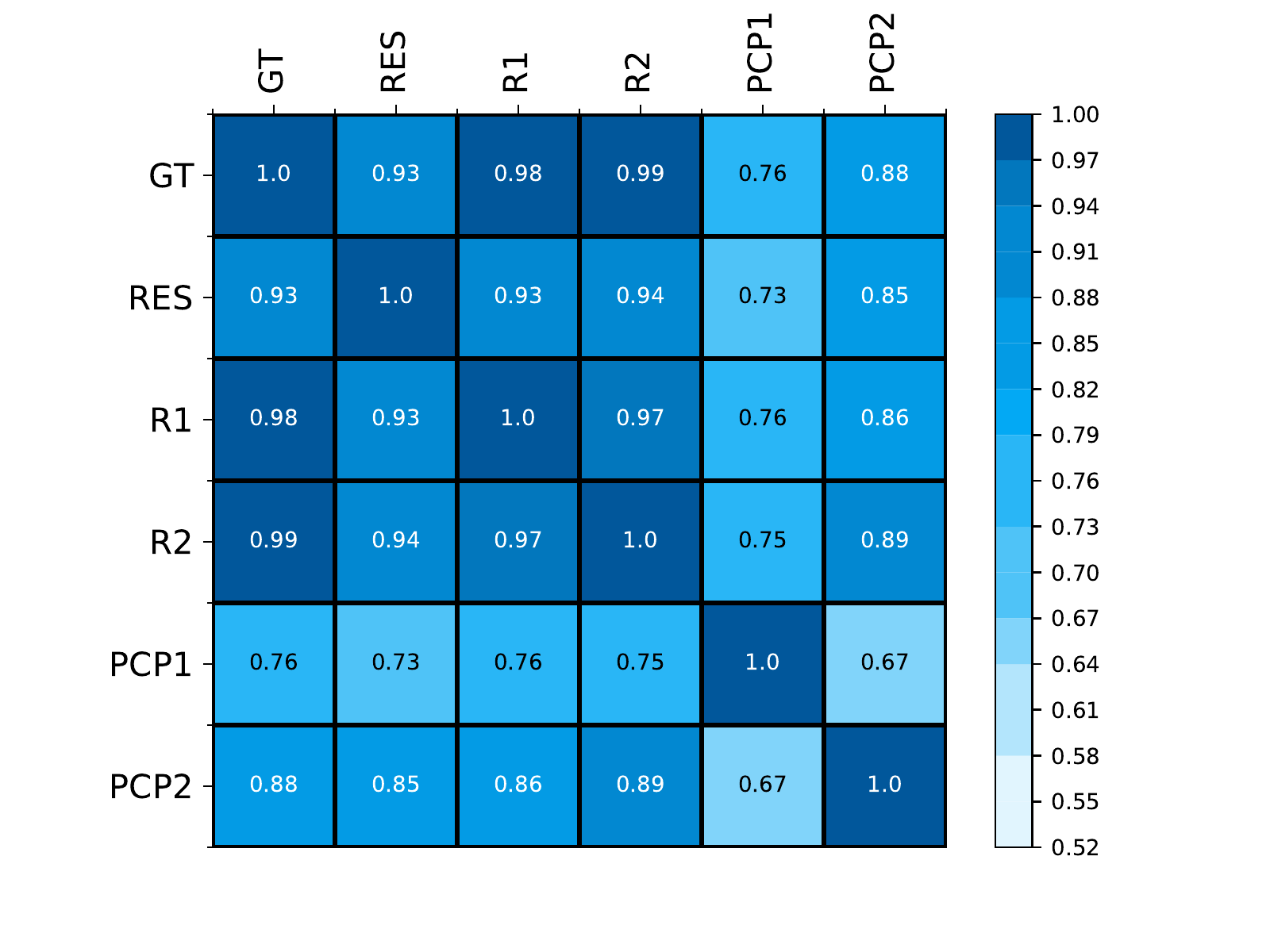}
    \caption{Cohen's Quadratic Kappa: all against all raters for Test Set \#1}
    \label{fig:suppli_ts1_kappa}
\end{figure}

\begin{figure}
    \centering
    \includegraphics{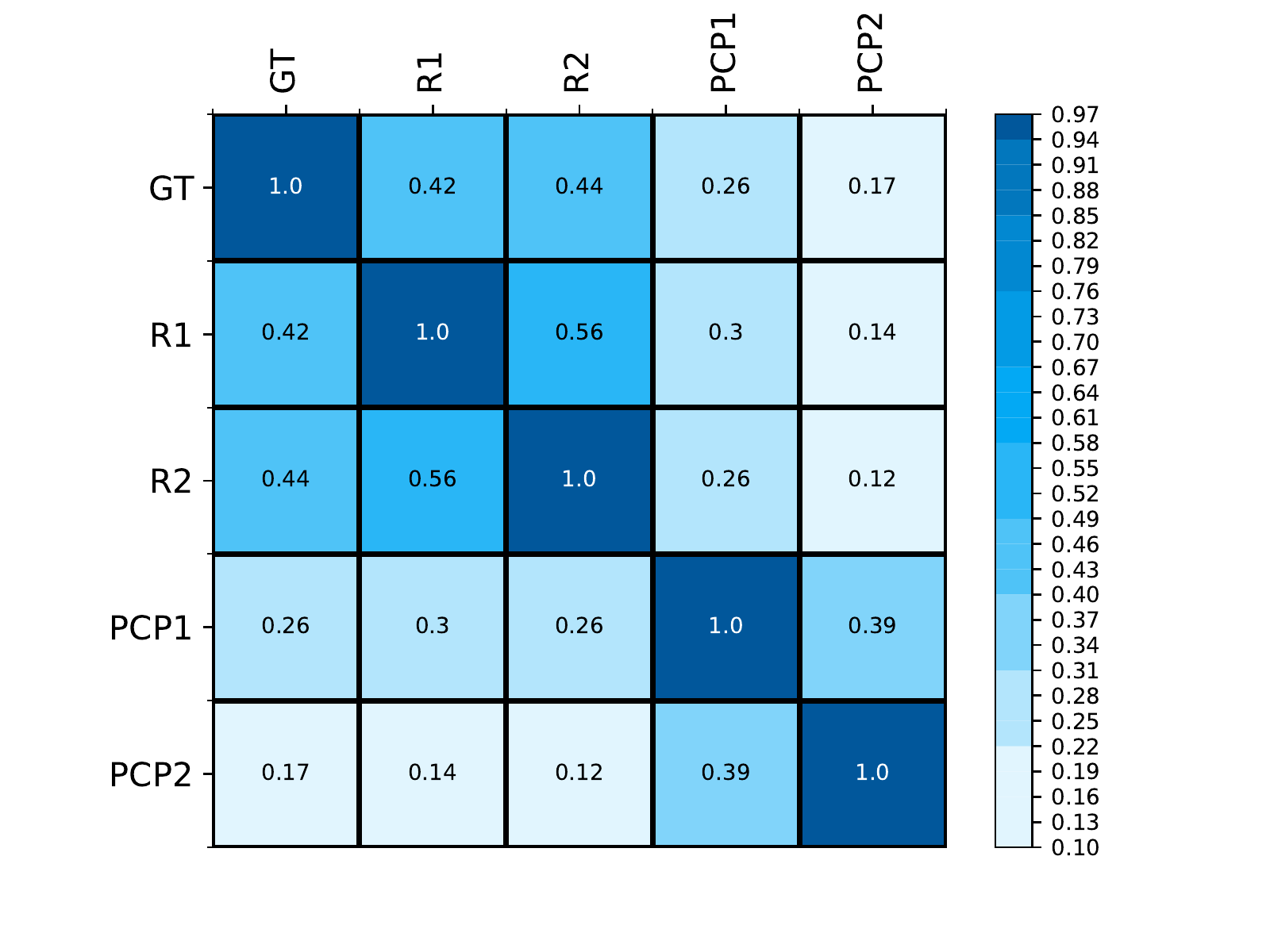}
    \caption{Cohen's Quadratic Kappa: all against all raters for Test Set \#2}
    \label{fig:suppli_ts2_kappa}
\end{figure}

\paragraph{Test Set \#1} \autoref{fig:kappa} shows inter-rater analysis using Cohen's Quadratic Kappa against the ground truth and the PCP1. \autoref{fig:suppli_ts1_kappa} and ~\autoref{fig:suppli_ts2_kappa} shows all against all agreement.


\section{Out-of-Distribution Experiment}
Initially, we assumed that there is a distribution shift between general population cases and challenging cases of wrist fracture. If true, this could indicate that the general population cases are in-distribution or in-domain data and challenging cases are out-of-distribution (OOD) data. Thereby, the performance could be improved if we would add the data from Test Set \#2 to the train set. 

Recent works on uncertainty estimation (for example by Lakshminarayanan et al. work~\cite{lakshminarayanan2017simple}) show that it is possible to detect OOD data samples using uncertainty estimates. To that end, we set up four Ensembles with 3, 5, 7, and 9 models, respectively. We did not use any cross validation for training Deep Ensemble, rather we split the whole training set into training and validation set and trained the Fracture Detection Block of our DeepWrist pipeline with different random initialization. Because of this, the approach Deep Ensemble lacked the ability to make use of transfer learning (as was done for the main model in the paper). We note that sole purpose of this experiment was to show that the challenging cases are not OOD data. The models for the ensembles were trained similarly to the main model shown in the paper, except, we did not use mixup.

\begin{figure}[ht!]
\centering
\begin{subfigure}{.5\textwidth}
  \centering
  \includegraphics[width=0.9\linewidth]{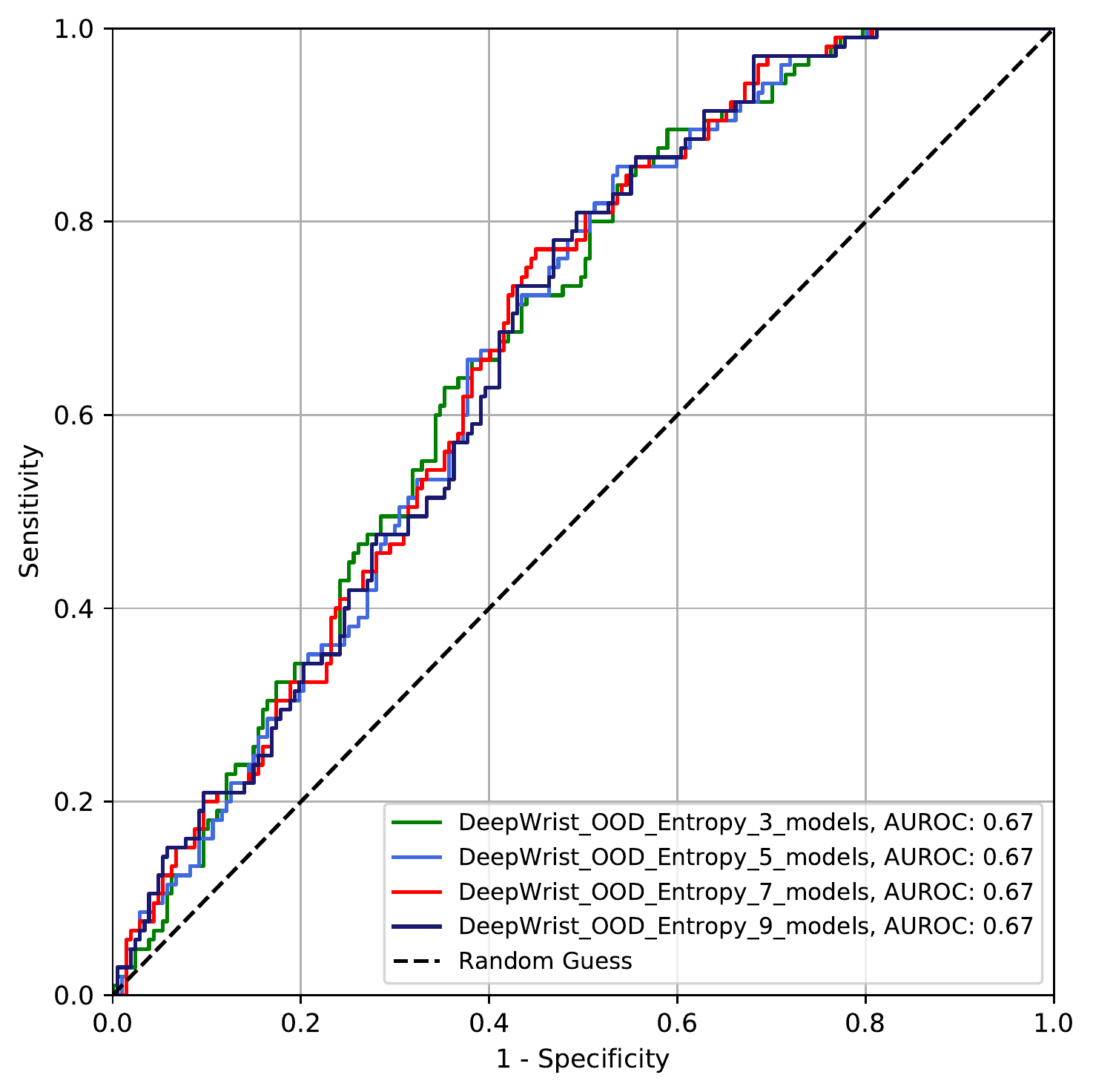}
  \caption{AUROC}
  \label{fig:roc_ood1}
\end{subfigure}%
\begin{subfigure}{.5\textwidth}
  \centering
  \includegraphics[width=0.9\linewidth]{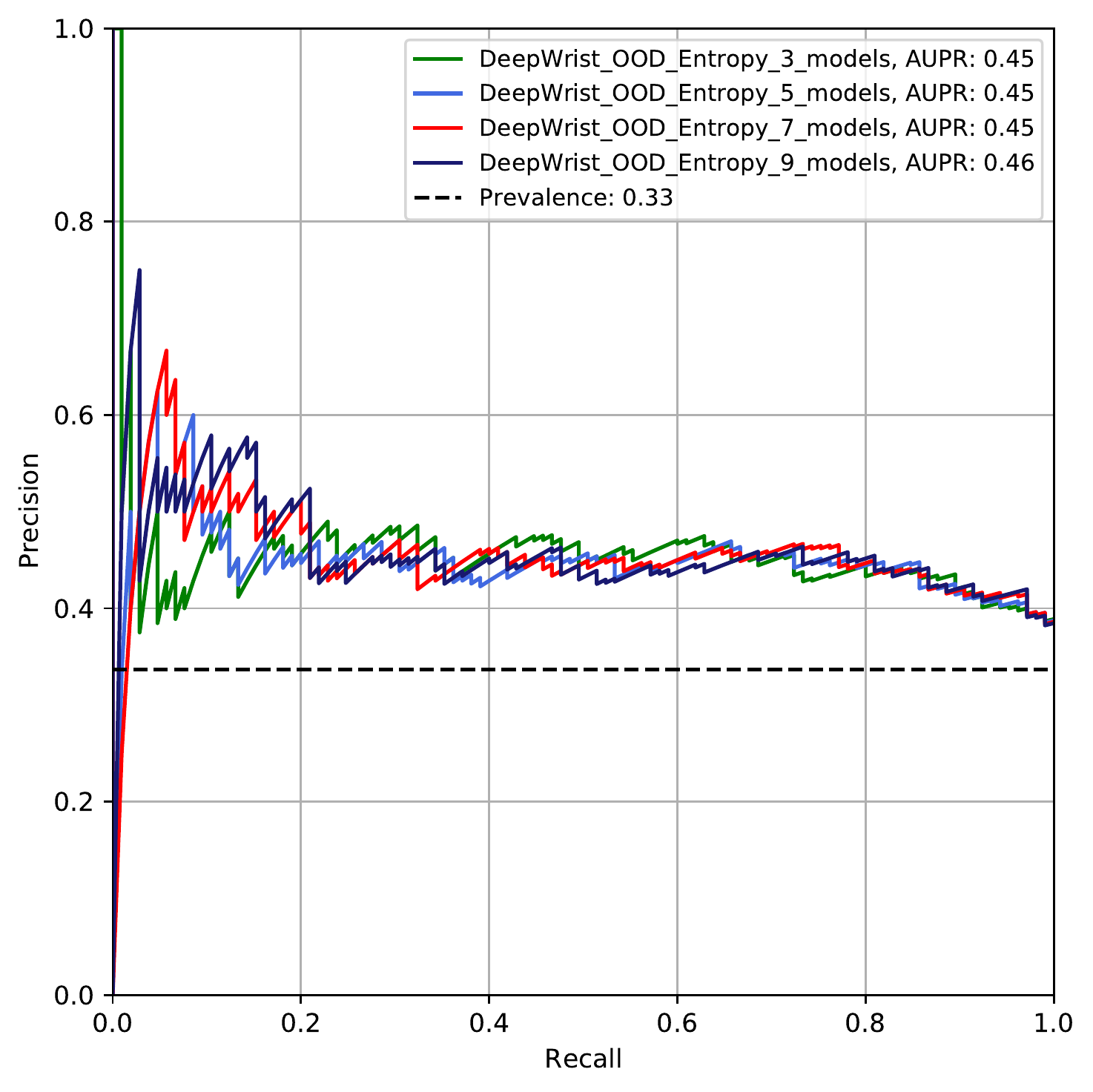}
  \caption{AUPR}
  \label{fig:pr_ood1}
\end{subfigure}
\caption{a) AUROC performance of OOD detection (by Entropy as uncertainty) using Deep Ensemble of 3,5,7 and 9 models respectively.  b) AUPR performance of OOD detection for the same Deep Ensemble. }
\label{fig:ood1}
\end{figure}

\begin{figure}[ht!]
\centering
\begin{subfigure}{.5\textwidth}
  \centering
  \includegraphics[width=0.9\linewidth]{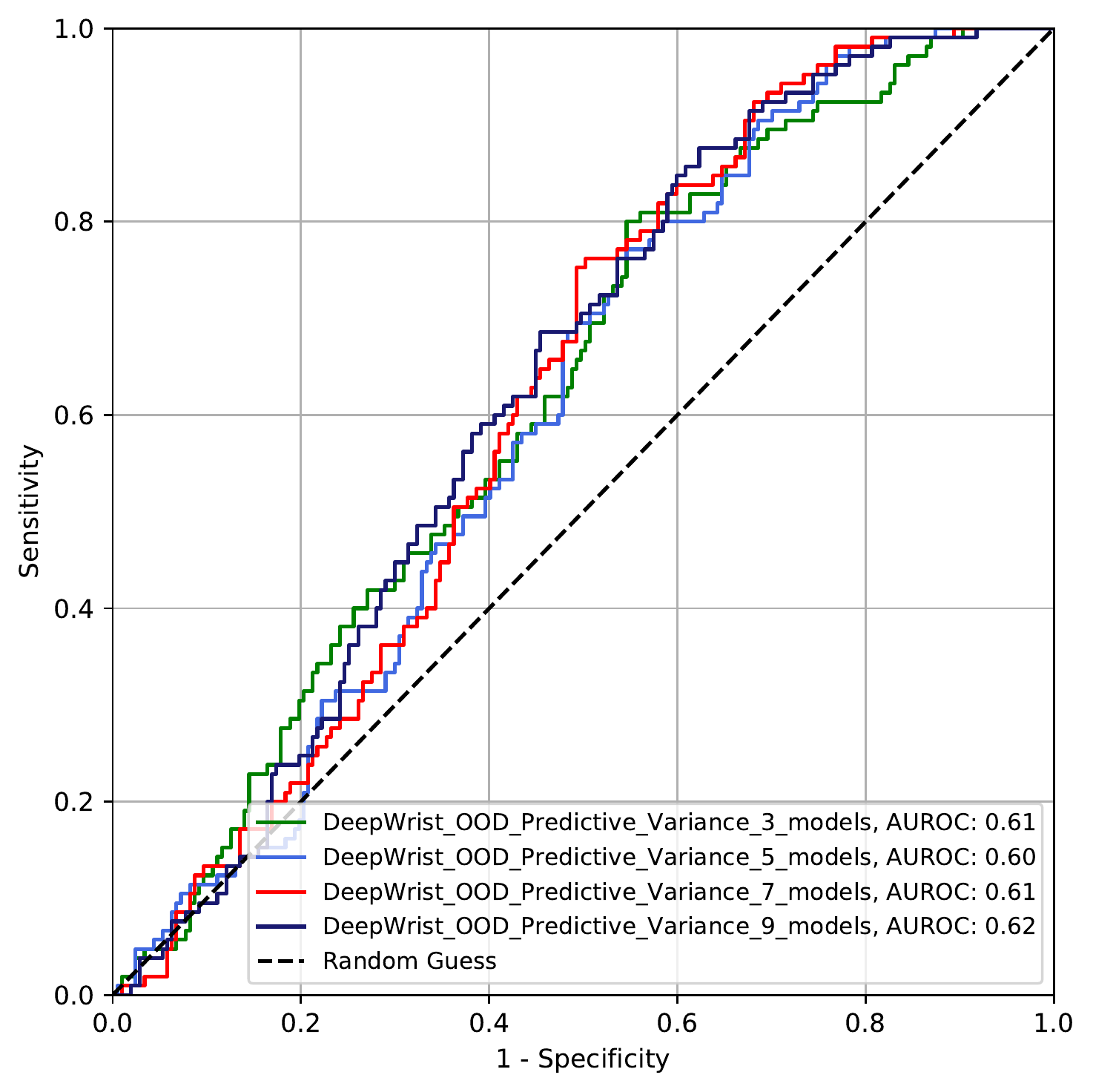}
  \caption{AUROC}
  \label{fig:roc_ood2}
\end{subfigure}%
\begin{subfigure}{.5\textwidth}
  \centering
  \includegraphics[width=0.9\linewidth]{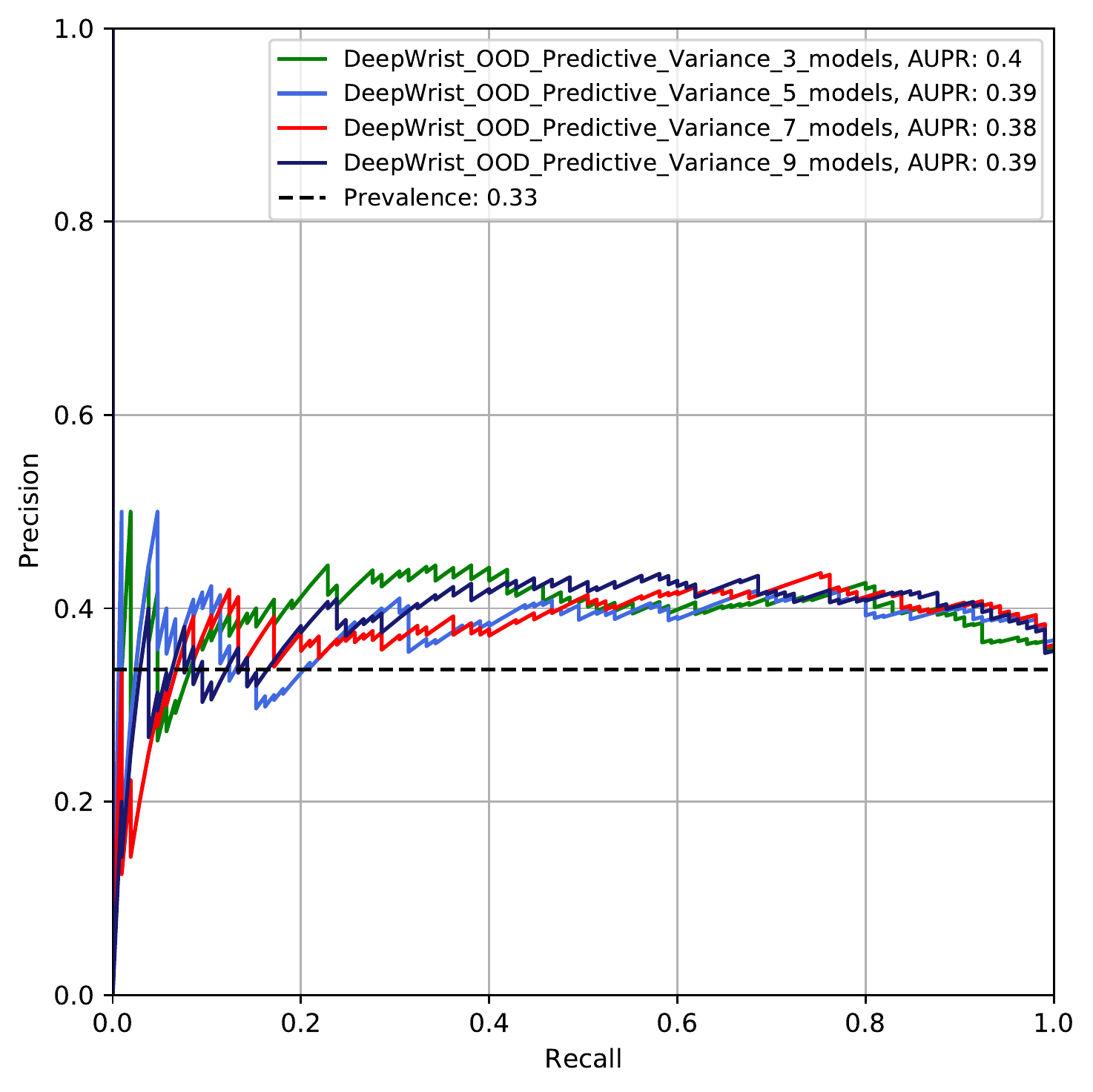}
  \caption{AUPR}
  \label{fig:pr_ood2}
\end{subfigure}
\caption{a) AUROC performance of OOD detection (by Predictive Variance as uncertainty) using Deep Ensemble of 3,5,7 and 9 models respectively.  b) AUPR performance of OOD detection for the same Deep Ensemble.}
\label{fig:ood2}
\end{figure}

\begin{figure}
    \centering
    \includegraphics[width=0.5\linewidth]{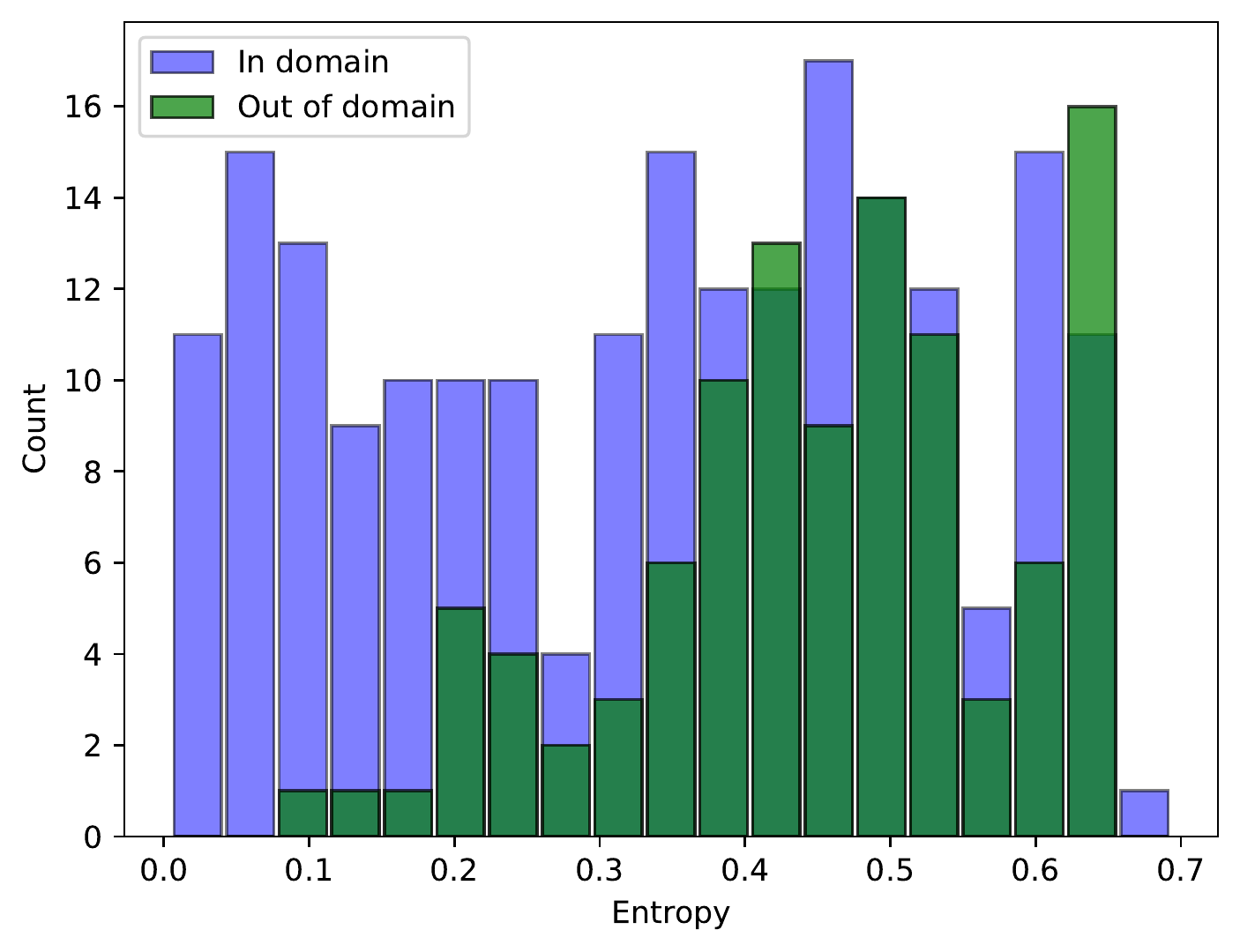}
    \caption{Entropy distribution of In-domain (general population cases) and Out-of-domain (challenging cases) data}
    \label{fig:hist_ent}
\end{figure}

We used Entropy and Predictive Variance as the estimated uncertainty of the corresponding prediction and used them to detect OOD samples. To obtain well calibrated uncertainty estimate, we calibrated the temperature of the models using the work of Guo et al.~\cite{guo2017calibration}.  In~\autoref{fig:ood1}, we show AUROC and AUPR performance of OOD detection with Entropy and in~\autoref{fig:ood2}, we show the same with Predictive Variance. It is evident from AUROC and AUPR that the OOD detection performance is poor. \autoref{tab:roc_ood} shows OOD detection AUROC with 95\% confidence interval for different ensemble settings. \autoref{fig:hist_ent} shows the entropy distribution of in-domain (general population cases) vs OOD (challenging cases) data. Clearly, there is no noticeable shift in these entropy distribution. Considering, all the AUROCs, AUPRs and the entropy distribution, we can conclude that the Deep Ensemble cannot differentiate between general population data and challenging data well.

\begin{table}[ht!]
\small
\centering
\begin{tabular}{ccc}
\toprule
\multirow{2}{*}[-0.2em]{\bfseries{\# models}} & \multicolumn{2}{c}{\bfseries{\multilinecell{AUROC for OOD Detection \\ (95\% CI)}}}\\
\cmidrule{2-3}
          & Entropy based             & Predictive Variance based\\ \midrule
3  &\multilinecell{0.67 \\ (0.61 - 0.73)}&\multilinecell{0.61 \\ (0.55 - 0.68)}\\
\midrule
5 & \multilinecell{0.67 \\ (0.61 - 0.73)}   &       \multilinecell{0.60 \\ (0.54 - 0.66)}  \\
\midrule
7 & \multilinecell{0.67 \\(0.61 - 0.73)}  &  \multilinecell{0.61 \\(0.55 - 0.67)}  \\
\midrule
9 & \multilinecell{0.67 \\ (0.61 - 0.73)} & 
\multilinecell{0.62 \\ (0.56 - 0.68)}  \\
\bottomrule

\end{tabular}
\caption{AUROC of Deep Ensemble for OOD detection}
\label{tab:roc_ood}
\end{table}

\begin{table}[ht!]
\small
\centering
\begin{tabular}{ccccc}
\toprule
\bfseries{Model} & \bfseries{LR} & \bfseries{Momentum} & \bfseries{Weight Decay} & \bfseries{Nesterov} \\
\midrule
SeresNet50 & $1e-1$, $1e-2$, $1e-3$, & 0.0, 0.5, 0.9 & 0.0, $1e-3$, $1e-4$, $3e-4$, & Yes, No \\
\midrule
Hourglass Net & $1e-1$, $1e-2$, $1e-3$, & 0.0, 0.5, 0.9 & 0.0, $1e-4$, & Yes, No \\
\bottomrule

\end{tabular}
\caption{Hyperparameters search space. We kept the optimizer fixed to SGD. Batch size was 32 for SeresNet50 and 24 for the hourglass model (KNEEL~\cite{tiulpin2019kneel})}
\label{tab:hyp}
\end{table}
\clearpage
\bibliography{supplement}

%% file: section/introduction.tex
\section*{Introduction}
Wrist fractures are the most common type of fractures~\cite{rundgren2020epidemiology} and typically indicate the fractures in the distal radius or ulna bones. The prevalence of wrist fractures is high, and according to the recent data, approximately 18 million hand and wrist fracture incidents occurred worldwide~\cite{crowe2020global}. Population-wise, $162$ cases of the distal radius or ulna fractures occur on average per $100,000$ inhabitants per year in the United States~\cite{karl2015epidemiology}. In the northern countries the incident rate is even higher. For example, in Finland the number of incidents is $258$ per $100,000$ inhabitants annually~\cite{flinkkila2011epidemiology}. 

Various types of treatments are available depending on the fracture's severity. Conservative casting and splinting are used for simple, acute, and nondisplaced fractures~\cite{knott2020casting}. Besides, a large number of patients are treated with operative treatment (surgery)~\cite{taljanovic2003fracture}. As an example from an economical point of view, Dutch Injury Surveillance System analysis shows that annual expenditure for wrist and hand injuries in the Netherlands is over \EUR{$540,000,000$}~\cite{de2012economic}. In addition to the financial burden, wrist fractures significantly reduce the quality of life. A study on Australian older adults shows that the loss in Health Related Quality of Life due to wrist fracture takes around 18 months for recovery~\cite{abimanyi2015changes}. Due to the aforementioned facts, wrist fractures pose a significant healthcare burden worldwide.

Conventional radiography (X-ray imaging) is used routinely as the first-line tool for wrist fractures diagnosis~\cite{basha2018does}. All plain radiographs are taken in certain projection views: lateral (LAT), posteroanterior (PA), anteroposterior (AP), or oblique. For most of the cases, X-ray imaging is sufficient to keep the high quality of care, and it emits substantially less radiation to the patients than volumetric modalities do, such as computed tomography (CT)~\cite{smith2009radiation}. 

Wrist X-ray images are usually taken in an emergency room and visually inspected by the attending physician, or if available, by a radiologist. Diagnostic errors, especially misdiagnosis of fractures, are common issues in the haste of the emergency setting~\cite{guly2001diagnostic}. Generally, the diagnostic performance of a physician can be affected by multiple factors, such as work overload, fatigue, and lack of experience~\cite{hallas2006errors, lindsey2018deep}. Many image interpretation errors could be avoided in the emergency room if the radiographs would be always instantly read by a radiologist or analyzed automatically providing support in the decision-making process. 

During recent years, Deep Learning (DL) has been widely applied in the realm of musculoskeletal radiology. In the domain of automatic fracture detection, DL has been used in application to radiographs on various body parts: ankle ~\cite{kitamura2019ankle}, hip~\cite{adams2019computer, badgeley2019deep,krogue2020automatic}, humerus~\cite{chung2018automated}, and wrist~\cite{bluthgen2020detection, thian2019convolutional, lindsey2018deep,kim2018artificial}. The wrist fracture detection performances in these studies were reported to be relatively high -- the Area Under the Receiver Operator Characteristics curve (AUROC) was of above or equal to $0.80$ on a test set. However, all these studies lack the validation of the methods on difficult fractures, which are challenging to diagnose without CT, and can only be diagnosed by a very experienced professional. We note that in clinical practice, CT is applied rather seldom, mostly in the cases where a fracture is clinically obvious or heavily suspected, but the radiographs do not show any signs of it~\cite{welling2008mdct}. Therefore, having a reliable diagnostic process for these rare cases directly impacts patient care, and if one wants to establish a fully automatic assessment of wrist images in a clinical setting, a special attention needs to be paid to the challenging cases.

Generally, rare clinical cases are rather unaddressed as a separate stratum in the state-of-the-art medical imaging studies as a result their effects go unnoticed due to hidden stratification issue~\cite{oakden2020hidden}. Recent studies on hidden stratification show that performance drop can be significantly high for an unaddressed stratum~\cite{oakden2020hidden, chedid2020synthesis}. Uncertain wrist cases that needed CT imaging form such stratum, and are of primary interest in this work.

\begin{figure}
\centering
\includegraphics[width=1.0\linewidth]{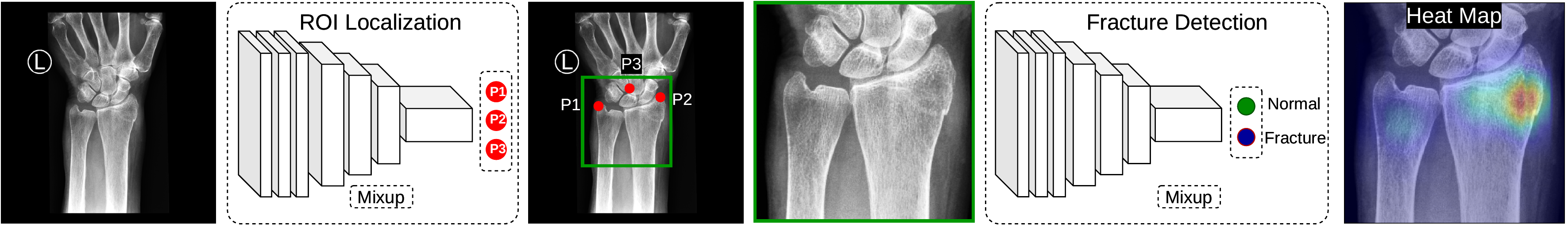}
\caption{DeepWrist Pipeline. From left to right: a wrist radiograph is passed to ROI (Region of Interest) localization block, which predicts three landmark points (P1, P2 and P3). Subsequently, these landmarks are used to crop an ROI image from the original radiograph. Finally, we utilize a fracture detection block which predicts whether the radiograph is normal or has a fracture. In addition to the prediction, we generate an explanation of the decision using a GradCAM technique.}
\label{fig:deepwrist}
\end{figure}

In this paper, we highlight the issue of hidden stratification in the realm of distal radius wrist fracture detection. In the sequel, we use the term wrist fractures for compactness, implying the fractures of the distal radius bone. The main contributions of our work can be summarized as follows:
\begin{itemize}
    \item We develop an open-source wrist fracture diagnosis method -- DeepWrist (see ~\autoref{fig:deepwrist}). This method is a two-stage pipeline, which utilizes anatomical landmark localization and image classification models, and reveals the local decision explanation using a GradCAM approach~\cite{selvaraju2017grad}. We show that on a general independent test set, this method yields high performance.
    \item For the first time in the realm of automatic wrist fracture detection, we show that a DL model trained on general population cases does not perform well on the difficult cases, which needed a CT imaging for diagnosis. 
    \item We show that despite a prior belief on domain shift between the general population and difficult cases, state-of-the-art techniques for estimating uncertainty in DL, such as Deep Ensembles~\cite{lakshminarayanan2017simple} are barely able to discriminate between these two sets of images.
    \item Finally, we compare the performance of our model and human physicians with various experience levels to investigate whether the aforementioned discrepancy is natural for them.
\end{itemize}

%% file: section/methods.tex
\section*{Materials and Methods}
\label{sec:method}

\subsection*{Data}
\label{subsec:dataset}
\paragraph{Overview} Our study leveraged three datasets, where one was used for training, and the other two for testing. These datasets consisted of referrals, PA and LAT images, and radiology reports. All the data were extracted from the Oulu University Hospital's (OUH) Picture Archiving and Communication System (PACS) and the Radiology Information System. We used pseudonymization to keep patients' identities protected. The project was approved by the Ethics Committee of Northern Ostrobothnia Hospital District (decision number: 126/2014), and the patients' informed consent requirement was waived due to the retrospective nature of this study. All methods of this research were performed in accordance with the Declaration of Helsinki. 

\paragraph{Training dataset}
To create the training set, we biased our data selection keeping the ratio of fractures $50\%$. Initially, our training dataset included $1000$ cases with distal radius fractures. Subsequently, images, which had artifacts (reasons -- non-diagnostic quality or implants) were removed leaving $953$ distal radius fracture cases. In total, $1946$ wrist studies ($3873$ PA and LAT images) were used in our training set. All the cases in this training set were the general fracture cases and it did not contain any challenging cases, for which an additional CT imaging was required. 

We annotated the training images based on the radiology reports: every image was visually inspected, and an existing radiology report was then manually labeled as \emph{normal} or \emph{fracture} (only distal radius fractures are considered) by a medical student who received basic training in diagnostic radiology. Thereby, we assigned the same label to both PA and LAT images. Detailed label and projection view distributions of all datasets are shown in~\autoref{tab:lab_dis}. The details on sex and age distribution can be found in Supplementary Section S1. 

\paragraph{Landmark localization data} As our pipeline leveraged two parts -- Region of Interest (ROI) localization block and fracture detection block, we had to perform the manual annotation for the ROI localization block. We annotated $3820$ out of $3873$ wrist radiographs from the training dataset with the anatomical landmarks (see~\autoref{fig:deepwrist}) using the VGG Image Annotator (VIA)~\cite{dutta2019vgg}. Here, $3056$ radiographs were used for training, and $764$ radiographs were used for measuring the accuracy of the ROI localization block. An analysis on intra-rater variability is discussed in Supplementary Section S2.

\input{tables/ds_table}

\paragraph{General population test set} The test set \#1 or the general population test set initially consisted of $210$ patient cases which were collected randomly from the Oulu University Hospital's PACS and did not require additional CT imaging for diagnosis. Three cases out of $210$ had implants, thus were excluded from the final analysis leaving $207$ cases with an equal number of PA and LAT radiographs where $129$ of the cases were annotated as fracture and $78$ as normal (see ~\autoref{tab:lab_dis} for details). All images in this set were acquired from the emergency department. We utilized an annotation strategy similar to that of training dataset and used radiology reports to create the initial labels for these general population data. The reports in this dataset were created by the total of $16$ radiology residents with work experience ranging from $16$ to $53$ months (median -- $35$ months).

Besides the annotations produced from the radiology reports, all the radiographs in this dataset were re-read by two board-certified radiologists independently without the knowledge of initial radiology reports. Radiologists were specifically asked to give a yes-or-no answer whether there is a fracture in the distal radius or not to keep the labeling in line with the training data. In case of disagreement (3 cases), a consensus decision was made. Consensus-based labels were used as the ground truth for the test set \#1. Beside the annotations from the board-certified radiologists, we included the annotations by other practitioners: the radiographs in this set were independently read by 2 primary care physicians with 3 and 4 years of clinical experience. 

\paragraph{Challenging test set} The test set \#2 or the test set of challenging cases had a total of $105$ patient cases. These data were deemed hard for diagnosis from X-ray images, thus the presence of fracture was determined by CT imaging. Among the extracted $105$ cases, $85$ cases were found normal and $20$ were found to have distal radius fracture from the radiology report (see ~\autoref{tab:lab_dis} for details). The annotations derived from the CT report were used as the ground truth for this dataset. The two board-certified radiologists and two primary care physicians, who annotated the test set \#1, also annotated the test set \#2. 

\subsection*{DeepWrist pipeline}
\label{subsec:overview}
\paragraph{Overview and experimental setup}
\autoref{fig:deepwrist} shows a graphical illustration of our approach. The whole pipeline comprises two parts -- ROI localization block by landmark localization and fracture detection block. The former part is based on the KNEEL method by Tiulpin \etal ~\cite{tiulpin2019kneel}, and it was trained to localize three anatomical landmarks (see ~\autoref{fig:deepwrist}). Using these landmarks, we cropped the ROI to include the part of the image that contains the distal radius bone. The latter part of our pipeline is a CNN based classifier, pre-trained on ImageNet dataset, and subsequently trained on our training dataset. 

All the experiments were conducted using PyTorch~\cite{paszke2019pytorch} with a PytorchLightning wrapper~\cite{falcon2019pytorch} for executing training and inference processes. SOLT~\cite{solt2019} library (version 0.1.8) was used for data augmentation. We ran all our experiments using a single Nvidia Geforce RTX 2080 Ti GPU. For each view (PA and LAT), separate ROI localization and fracture detection blocks were trained. 

Except for the final testing, all the experiments were conducted using cross-validation (CV) to determine the best hyperparameters. The classifiers' thresholds and the temperature hyperparameters of Deep Ensemble were maximized in an out-of-fold cross-validation setting. Supplementary Table S3 shows the settings used for hyperparameters selection. We used a 5-Fold CV to train the ROI localization block. To train the fracture detection block we also used the similar procedure. Here, we used the patient ID for group splitting to ensure that training and validation datasets did not intersect.

\paragraph{Pre-processing and augmentation} 
All the data were pre-processed before passing them through any of the blocks. After reading each radiograph, we used the global contrast normalization with initial clipping between the $5^{th}$ and $99^{th}$ intensity percentiles. 

Due to the images being of large size, we used bi-linear interpolation, and re-scaled the images to a lower pixel-spacing. Specifically, we used the target pixel spacing of $0.27mm$ for the PA view and $0.35mm$ for the LAT view, to train the fracture detection block. For the ROI localization block, pixel spacing was not fixed, rather it was dependent on the expected size of input to the block in pixels which was $256 \times 256$.

For training, we used heavy data augmentations. We applied cutout~\cite{devries2017improved}, jittering, random color padding on a particular side, downscaling, flipping, rotation, shearing, padding, salt and pepper, blur, noise and gamma correction for the ROI localization block. For the fracture detection block we used similar augmentations. More details about the data augmentations are shown in the source code.

During inference, we did not use any augmentation for ROI localization block but we used Test-Time Augmentation (TTA) for fracture detection block to improve the performance. For the TTA, we used gray scale to color conversion, flipping and five-crop on both flipped and unflipped images. 

\paragraph{ROI localization block}
\label{subsec:training}
This module of the pipeline is a landmark localizer, which learns to identify three major key points in the wrist radiographs. After localization, we crop the ROI using the detected landmark points.

For PA view the landmarks were placed at the top of distal ulna, top of distal radius and the center of the wrist (see~\autoref{fig:deepwrist}). For the LAT view, the landmarks were two distinguishable points on two sides of the top part of radio-ulna, and the center of wrist. 

In short, our landmark localizer uses an hourglass network~\cite{newell2016stacked}, with a soft-argmax layer to predict the landmark coordinates directly. We utilized the existing method and the open-source codebase from the KNEEL method~\cite{tiulpin2019kneel}. To train this model, we used a Stochastic Gradient Descent(SGD) optimizer with a learning rate of $1e-1$ with no momentum and a batch size of $24$. The localization pipeline was trained for $300$ epochs with a learning rate drop at $150^{th}$, $200^{th}$ and $250^{th}$ epochs by a factor of $10$. 

Since, ROI localization is a crucial part of our fracture detection pipeline, one has to ensure that the absence of failures on the datasets. Thus, to regularize the training, we used \textit{mixup}~\cite{zhang2017mixup}. Such strategy had shown to improve adversarial robustness, and improve generalization of Deep Neural Networks~\cite{zhang2017mixup}. We also observed similar effects in our cross-validation experiments. 

Briefly, \textit{mixup} aims to convexify the training set by creating interpolated samples:
\begin{align}
    x_{mix} = \lambda x_1 + (1-\lambda)x_2\\
    y_{mix} = \lambda y_1 + (1-\lambda)y_2
\end{align}
where $\lambda \sim \textrm{Beta}(\alpha, \alpha)$. Empirically, we found that training with $\alpha = 0.4$ works the best with our data, and as recommended by the authors of KNEEL~\cite{tiulpin2019kneel}, we did not use the weight decay. 

As mentioned earlier, we used the generated landmarks to create the ROI for fracture detection block. For that, we computed the center of mass of the landmark coordinates and added a top padding to the obtained coordinate point to calculate the center for cropping the ROI from the original DICOM image. In our experiments, the PA ROI had a size of $70mm \times 70mm $ with a $15mm$ top padding and LAT ROI has a size of $90mm \times 90mm$ with a $20mm$ top padding. These values for cropping the ROI were chosen empirically based on the visual inspection on CV. As mentioned earlier, we had 5 models from 5-fold CV. During inference, we formed a 5-model ensemble, averaged the predicted landmarks coordinates from five models and used them as the predicted landmark coordinates of the block. After the ROI localization block was trained, we applied it to generate the ROIs for the whole training dataset to train the fracture detection block.

\paragraph{Fracture detection block}
\label{subsec:fd}
We used a SeresNet50~\cite{hu2018squeeze} model pre-trained on ImageNet~\cite{imagenet_cvpr09} dataset for fracture detection block. We added a \emph{dropout} layer with 50\% probability before the \emph{fully connected} layer of the network (randomly reset to predict two classes, contrary to 1000 classes in ImageNet). The remaining part of the model architecture was taken from the work by Hu \textit{et al.}~\cite{hu2018squeeze}. Similar to the ROI localization block, we used an SGD optimizer with a learning rate of $1e-1$, batch size of $32$ and a weight decay of $1e-4$. We did not use any momentum for the training. The model was trained for $300$ epochs with a learning rate drop at $150
^{th}, 200^{th}, 250^{th}$ epochs by a factor of $10$. For the first $10$ epochs, we only trained the classifier part of the SeresNet50 and after that for the rest of the remaining epochs, we trained the full network.

\paragraph{Multi-view ensembling}
To leverage the radiographs from both PA and LAT views, we created an Ensemble, which computed the average of the underlying blocks' predictions (5 models from each CV fold). We note that in the case of fracture detection block, we applied TTA to each individual item in the Ensemble before averaging. The whole prediction strategy is visualized in Supplementary Figure S1. 

\paragraph{Evaluation of distribution shift}
To rule out the possibility that the hard cases have a distribution shift from the distribution of general population cases which is negatively affecting the performance of fracture detection for hard cases, we conducted experiments using Deep Ensemble~\cite{lakshminarayanan2017simple} approach to detect hard cases as out-of-distribution (OOD) data. We trained the above described fracture detection block, but without transfer learning, to ensure diversity in coverage of parameters' posterior distribution modes. For details about this experiment, see Supplementary Section S4. 

\subsection*{Results interpretation}
\paragraph{Decision explanation via GradCAM}
To interpret the predictions of the fracture detection block, our pipeline produces a heat map focusing the part of radiograph, which positively affected the outcome of the model. For this, we used GradCAM ~\cite{selvaraju2017grad} technique. In brief, GradCAM computes a weighted sum of the feature maps in the penultimate layer of the neural network. The weights for this summation are obtained by back-propagating the decision of choice (fracture in our case). 

\paragraph{Metrics and statistical analyses}
We used multiple metrics to interpret the results. In our notation, positive cases indicate fractures and negative indicates -- their absence. We assessed the performance of the fracture detection block as the total performance of our pipeline. The main metrics were the AUROC and Area Under Precision-Recall Curve (AUPR). Using these two metrics in conjunction is important, as the label distribution of test set \#2 is imbalanced (see ~\autoref{tab:lab_dis}). Apart from the metrics common in the machine learning literature, we also reported the metrics utilized by medical community -- Sensitivity (also known as Recall or True Positive Rate), Specificity (also known as Selectivity or True Negative Rate), Precision (also known as Positive Predictive Value), $F_1$ Score and Balanced Accuracy. Beside these metrics, we also used the Cohen's quadratic kappa ($\kappa$) for the inter-rater analysis. Kappa measures the agreement between two raters for the same cases.

As the aforementioned metrics are not suitable to assess the anatomical landmarks prediction quality, we used the Euclidean distance between predicted landmark coordinates and ground truth. Here, we defined different precision thresholds and calculated the percentage of correctly classified key points within $1mm$, $1.5mm$ etc.

To analyse the statistical significance, we used the stratified bootstrapping to compute the Confidence Interval (CI) of all the statistical metrics with $5,000$ iterations. We also used a logistic regression to assess the added value of our model to the confounding factors, such as age and sex on the test sets. We used \emph{statsmodels}~\cite{seabold2010statsmodels} for calculating the $p-value$.

%% file: tables/ds_table.tex
\begin{table}[ht!]
\small
\centering
\begin{tabular}{cccccccc}
\toprule
\bfseries{Dataset} & \bfseries{\# Cases} & \bfseries{\multilinecell{\# Fracture \\ Cases}} & \bfseries{\multilinecell{\# Normal \\ Cases}} & \bfseries{View} & \bfseries{\# Radiographs} & \bfseries{\multilinecell{\# Fracture \\ Radiographs}} & \bfseries{\multilinecell{\# Normal \\ Radiographs}} \\ \midrule
\multirow{2}{*}[-0.2em]{Training set} & \multirow{2}{*}[-0.2em]{1946} & \multirow{2}{*}[-0.2em]{953} & \multirow{2}{*}[-0.2em]{993} & PA & 1962  & 954  & 1008 \\ \cmidrule{5-8}
 & & & & LAT & 1911  & 946  & 965  \\ \midrule
 \multirow{2}{*}[-0.2em]{Test set \#1} & \multirow{2}{*}[-0.2em]{207} & \multirow{2}{*}[-0.2em]{129}& \multirow{2}{*}[-0.2em]{78}& PA  & 207  & 129 & 78 \\ \cmidrule{5-8} 
 & & & & LAT & 207 & 129  & 78 \\ \midrule
 \multirow{2}{*}[-0.2em]{Test set \#2} & \multirow{2}{*}[-0.2em]{105} & \multirow{2}{*}[-0.2em]{20}& \multirow{2}{*}[-0.2em]{85} & PA  & 105  & 20  & 85 \\ \cmidrule{5-8} 
 & & & & LAT & 105  & 20  & 85 \\ \bottomrule
 \end{tabular}
\caption{Datasets used in this study.}
\label{tab:lab_dis}
\end{table}

%% file: section/result.tex
\section*{Results}
\subsection*{Localization of anatomical landmarks}
 We analyzed the predictive performance of the landmark localizer as the predictive performance of ROI localization block. The landmarks are coarsely annotated for its training set as we do not need fine grained landmark coordinates for cutting a good ROI image. As a result the accuracy of landmark localizer (ROI localization block) is also evaluated with relaxation and tolerance. This block scores $0.70 \, (0.67 - 0.73)$ recall at $3mm$ precision, $0.88 \, (0.86 - 0.80)$ recall at $4mm$ precision and $0.96 \, (0.95 - 0.97)$ recall at $5mm$ precision on the holdout test set. We found this accuracy sufficient for ROI localization due to the subsequent cropping strategy which was also confirmed by the visual inspection on the out-of-fold validation data. Therefore we did not aim to further improve this block of our method. A more detailed evaluation of the ROI localization is presented in Supplementary Section S2.

\subsection*{Fracture detection}
\paragraph{Cross-validation and threshold optimization}
The out-of-fold validation accuracy was $0.95\,(0.93 - 0.97)$ and $0.98\,(0.97 - 0.99)$ for the PA and LAT views respectively. After training the models, we used validation predictions from all folds to identify the cut-off or threshold values. We found that $F_1$ Score was maximized when the probability threshold was of $0.41$ for the PA view and of $0.58$ for LAT view. For the final ensemble we used the average of these two thresholds ($0.5$). 

To decide whether the mixup~\cite{zhang2017mixup} technique would be used for this block, we trained this block with mixup ($\alpha=0.7$) and without mixup and evaluated them on the out-of-fold validation data. We found out that mixup slightly improves the performance on out-of-fold validation data therefore we kept the mixup technique for this block. 

\paragraph{Inter-rater agreement} Besides fracture detection performance we also analyzed inter-rater agreement among the human raters. We used Cohen's Quadratic Kappa for this purpose. The details of the inter-rater analyses can be found in Supplementary Section S3. 

For test set \#1, radiologist 2 had the most agreement with the consensus-based ground truth. Unlike radiologists, primary care physicians are not well trained on how to detect fracture accurately from plain radiographs which is reflected from the $\kappa$ values of two primary care physicians ($0.76$ and $0.88$) and two radiologists ($0.98$ and $0.99$) with respect to consensus. The radiology resident's $\kappa (0.93$) lay between the primary care physicians (PCP1 and PCP2) and the radiologists (R1 and R2). It is notable that primary care physicians disagreed between themselves the most. In fact, the PCP1 and PCP2 had the worst agreement ($\kappa$ = $0.67$ ) among all the raters. For test set \#2, all the raters have low agreement with the ground truth from the CT compared to the similar analyses in the test set \#1.

\input{tables/performance}

\subsection*{Test set \#1: general population test set}
For the test set \#1 the AUROCs were $0.98\,(0.97 - 0.99)$, $0.98\,(0.97 - 0.99)$ and $0.99\,(0.98 -0.99)$ for PA view, LAT view and Ensemble respectively (see ~\autoref{tab:test_perf}). In ~\autoref{fig:auroc}, we visualize the ROC curve for the test set \#1 along with the performance of radiology resident, two radiologists and two primary care physicians. In terms of sensitivity and specificity, the radiologists and the resident performed better than our pipeline. But the primary care physicians had mixed scores: PCP1 scored a lower specificity but a higher sensitivity and PCP2 scored a higher specificity but a lower sensitivity than our pipeline's corresponding score (see ~\autoref{fig:auroc} and ~\autoref{tab:acc} for details). 

The AUPR on test set \#1 is $0.99$ for all views and the Ensemble. In ~\autoref{fig:aupr}, we visualize the Precision-Recall curve along with the performance of other raters where the radiologists and resident performed better than the pipeline in terms of precision and recall. But like before, the primary care physicians had mixed scores: PCP1 scored a higher recall but a lower precision, and PCP2 scored a lower recall but a higher precision than our pipeline's corresponding score (see \autoref{fig:aupr} and ~\autoref{tab:acc} for details). Both AUROC and AUPR indicate that DeepWrist is a near-perfect classifier.

\subsection*{Test set \#2: hard cases}
\begin{figure*}[ht!]
\centering
\subfloat[AUROC on test set \#1 or trivial cases \label{fig:roc_ts1}]{\includegraphics[width=0.45\linewidth]{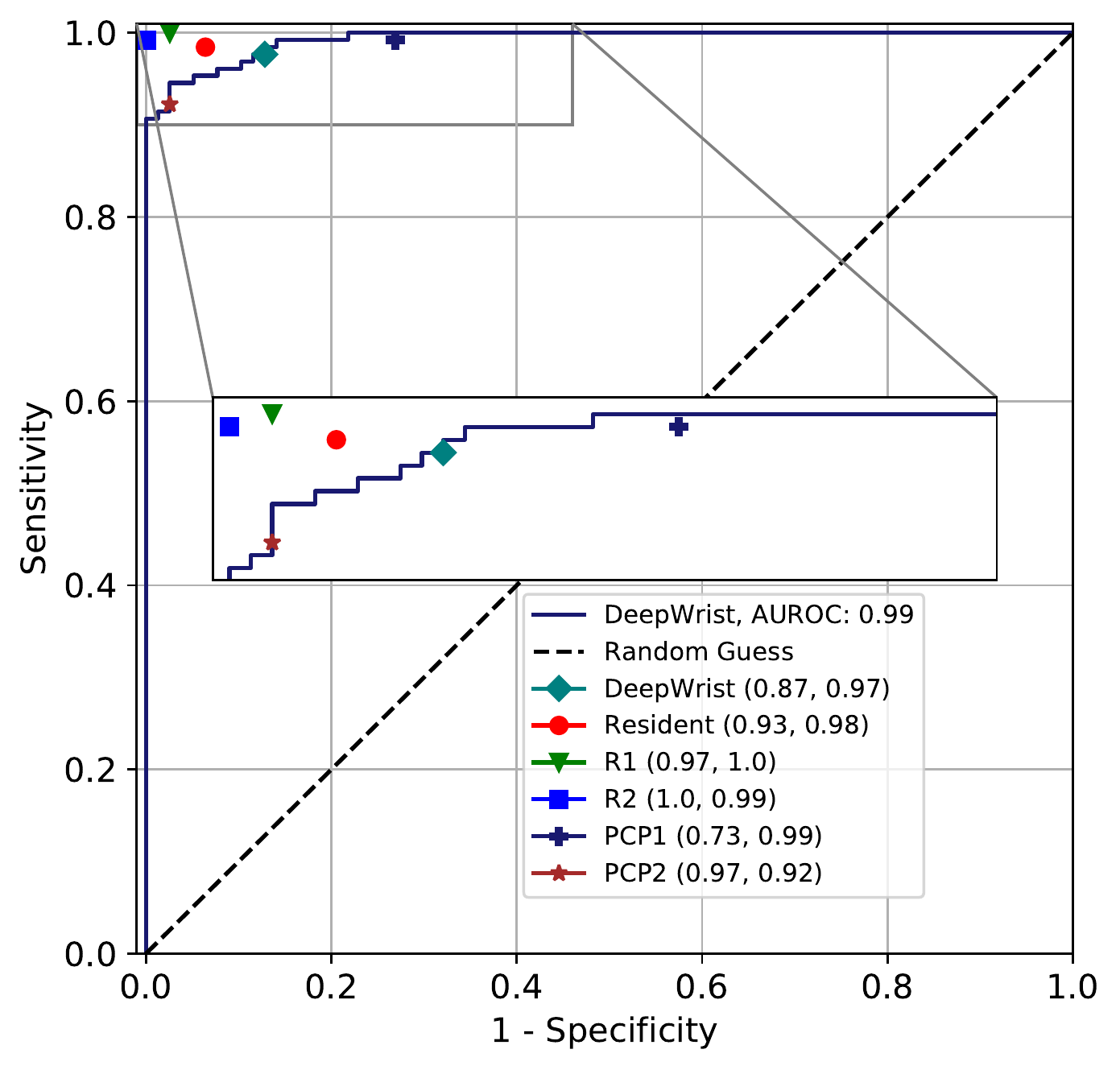}}
\hfil
\subfloat[AUROC on test set \#2 or hard cases    \label{fig:roc_ts2}]{\includegraphics[width=0.45\linewidth]{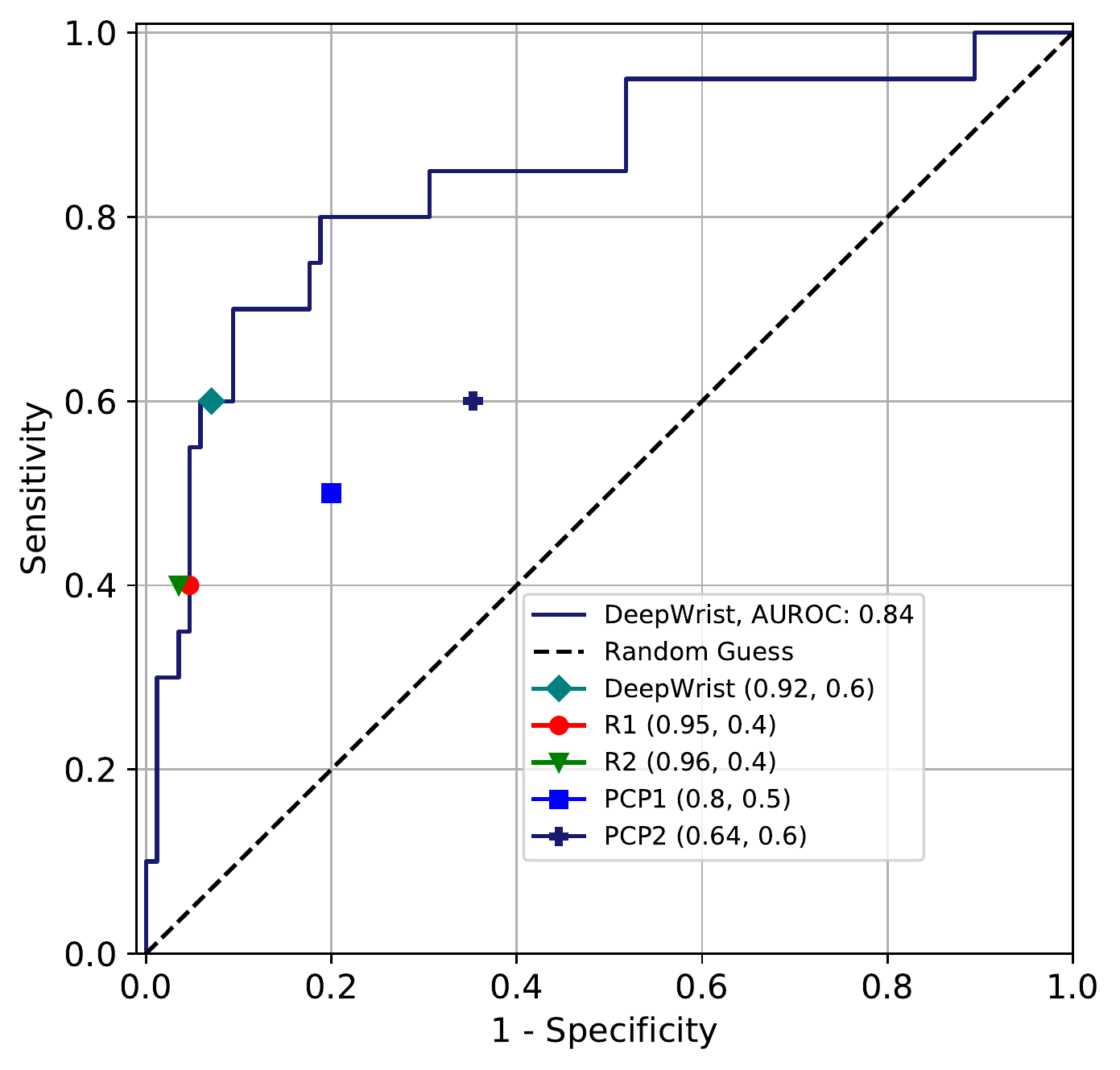}}

\caption{a) AUROC performance of DeepWrist on test set \#1 compared to a radiology resident, two radiologists (R1 \& R2), and two primary care physicians (PCP1, PCP2), b) AUROC performance of DeepWrist on test set \#2}
\label{fig:auroc}
\end{figure*}

\begin{figure*}[ht!]
\centering
\subfloat[AUPR on test set \#1 or trivial cases \label{fig:pr_ts1}]{\includegraphics[width=0.45\linewidth]{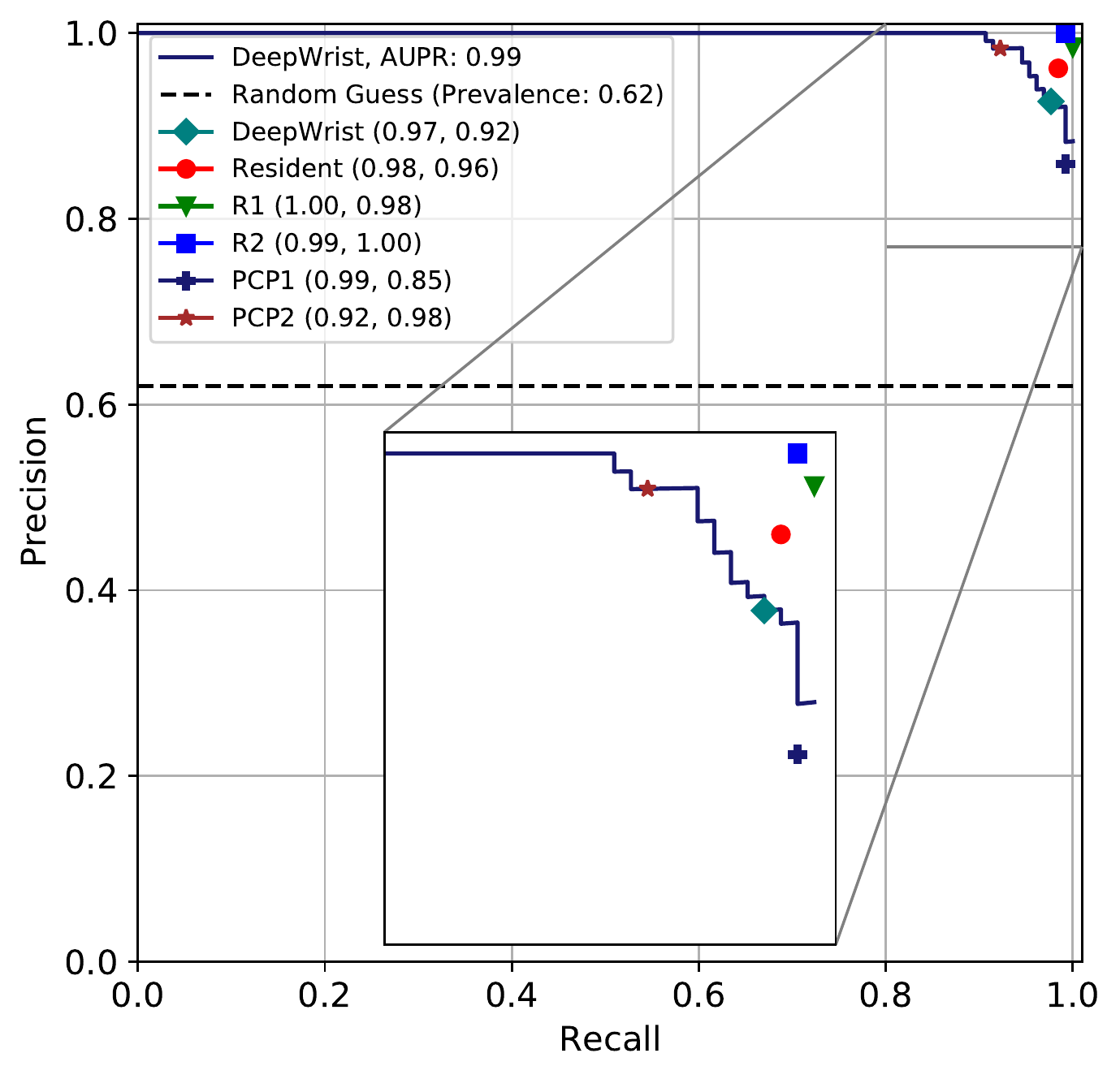}}
\hfil
\subfloat[AUPR on test set \#2 or hard cases \label{fig:pr_ts2}]{\includegraphics[width=0.45\linewidth]{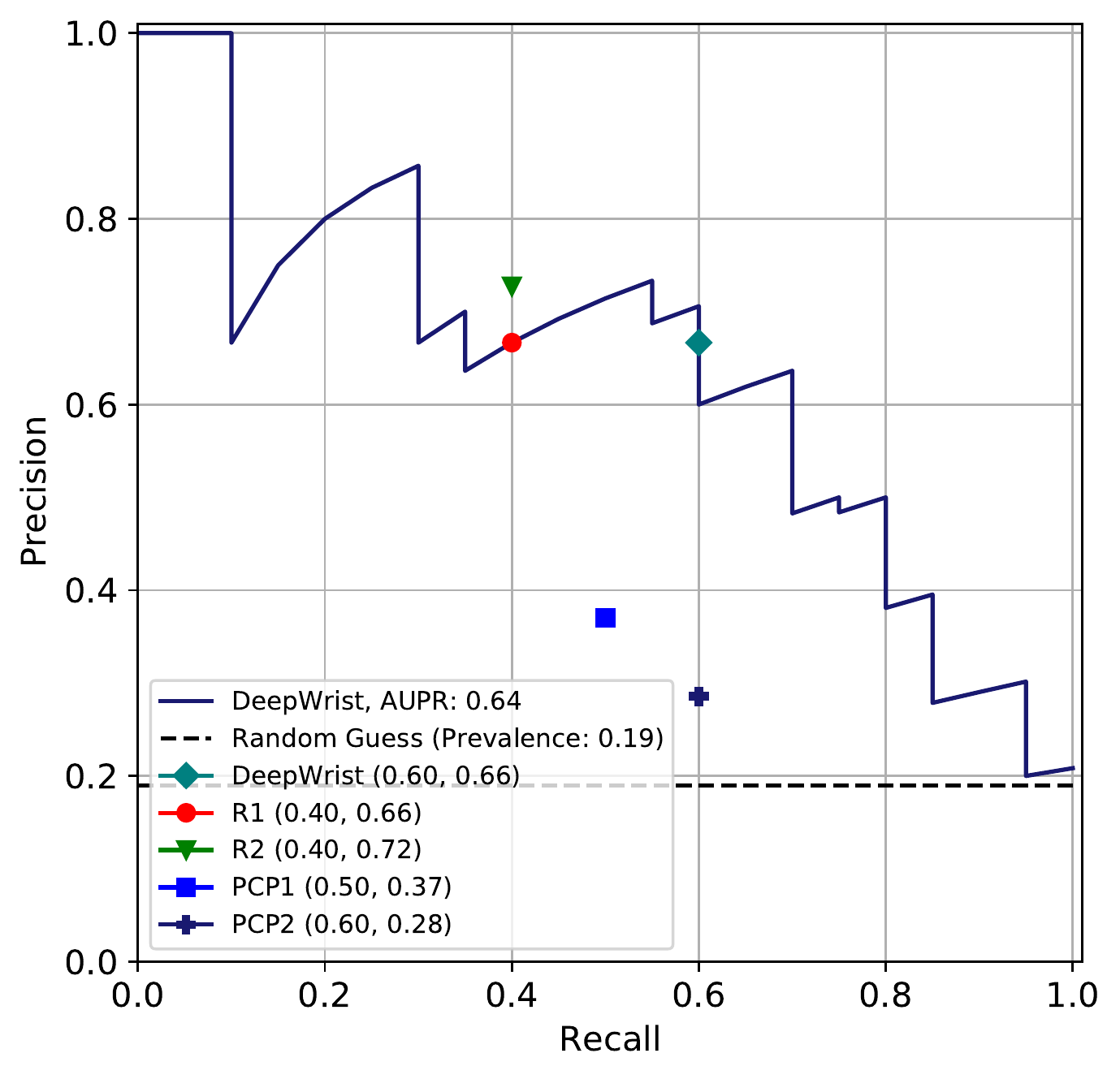}}

\caption{a) AUPR performance on the test set \#1 for DeepWrist, radiology resident, two radiologists (R1 \& R2), and two primary care physicians (PCP1 \& PCP2), b) AUPR performance of DeepWrist and other graders on the test set \#2. The plot highlights the drop in performance for both -- human raters of different expertise, and our method.}
\label{fig:aupr}
\end{figure*}

The AUROCs for the hard test set or test set \#2 were of $0.81\,(0.69 -0.91)$, $0.83\,(0.70 - 0.93)$ and $0.84\,(0.72 -0.93)$ for the PA view, LAT view and Ensemble respectively (see ~\autoref{tab:test_perf}). In subplot b) of ~\autoref{fig:auroc}, we show the performance of DeepWrist in terms of the sensitivity and specificity. Evidently the shown results are substantially lower compared to the results of test set \#1. The PR curve (\autoref{fig:aupr}) also indicates the same findings. We note that human raters also showed the drop in performance (see~\autoref{tab:acc_ts2}).

\input{tables/easy_set_stat}
\input{tables/hard_set_acc}

\paragraph{Analysis of pitfalls}
To analyse the pitfalls, we evaluated the impact of confounding factors (age and sex) using Logistic Regression to the predictions of our model. We found that for the test set \#1, age and sex are significantly associated with the outcome ($p<0.05$) but our model had also significant contributions ($p <0.001$). However, for the test set \#2 (hard cases), the $p$-value for DeepWrist was $0.43$, indicating that our method did not contribute to the outcome more than the confounding factors did.

\begin{figure}[hbt!]
\centering
\subfloat[   \label{fig:gc_pa}]{\includegraphics[width=.45\linewidth]{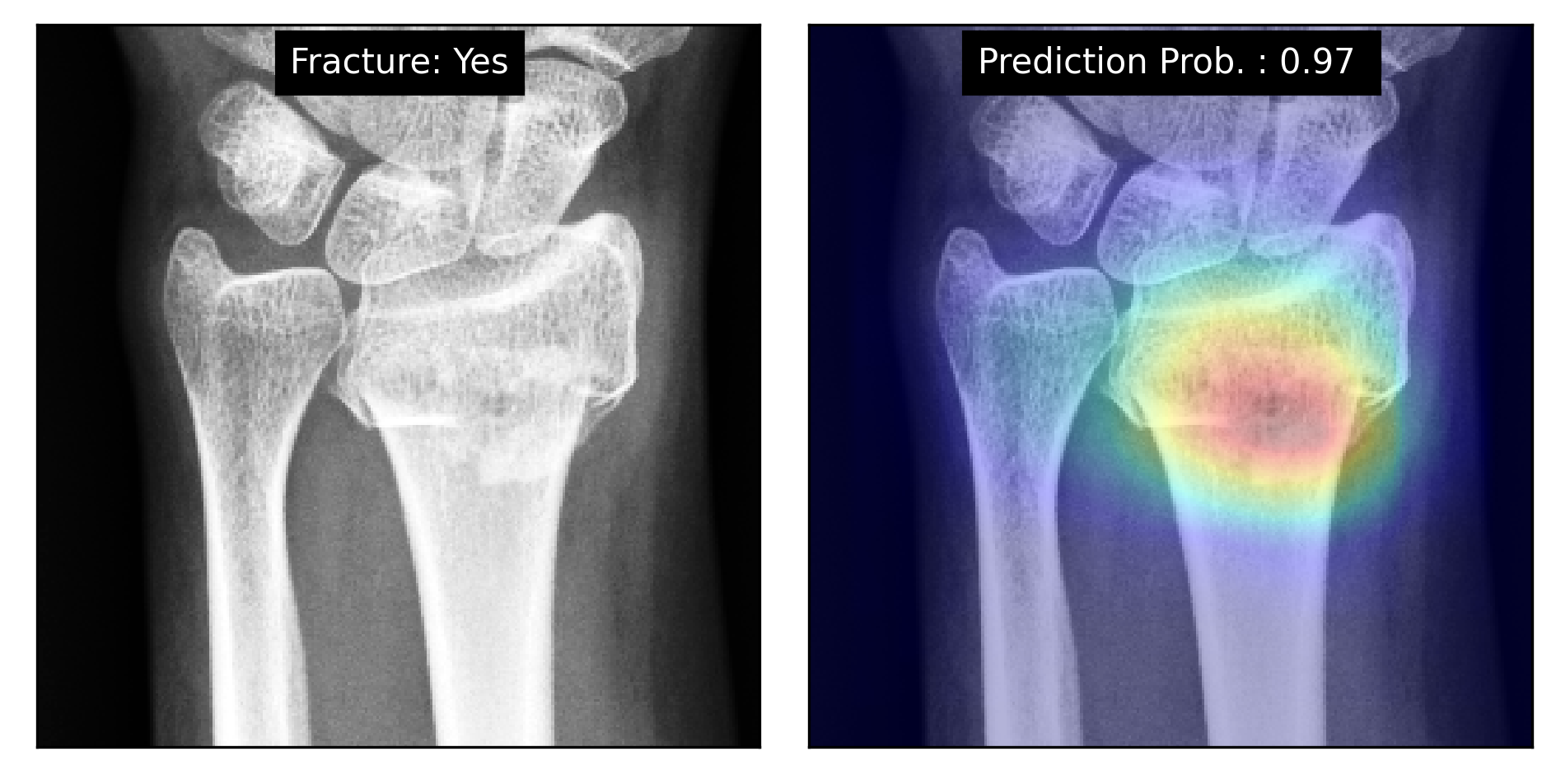}}
\hfil
\subfloat[ \label{fig_second_case}]{\includegraphics[width=.45\linewidth]{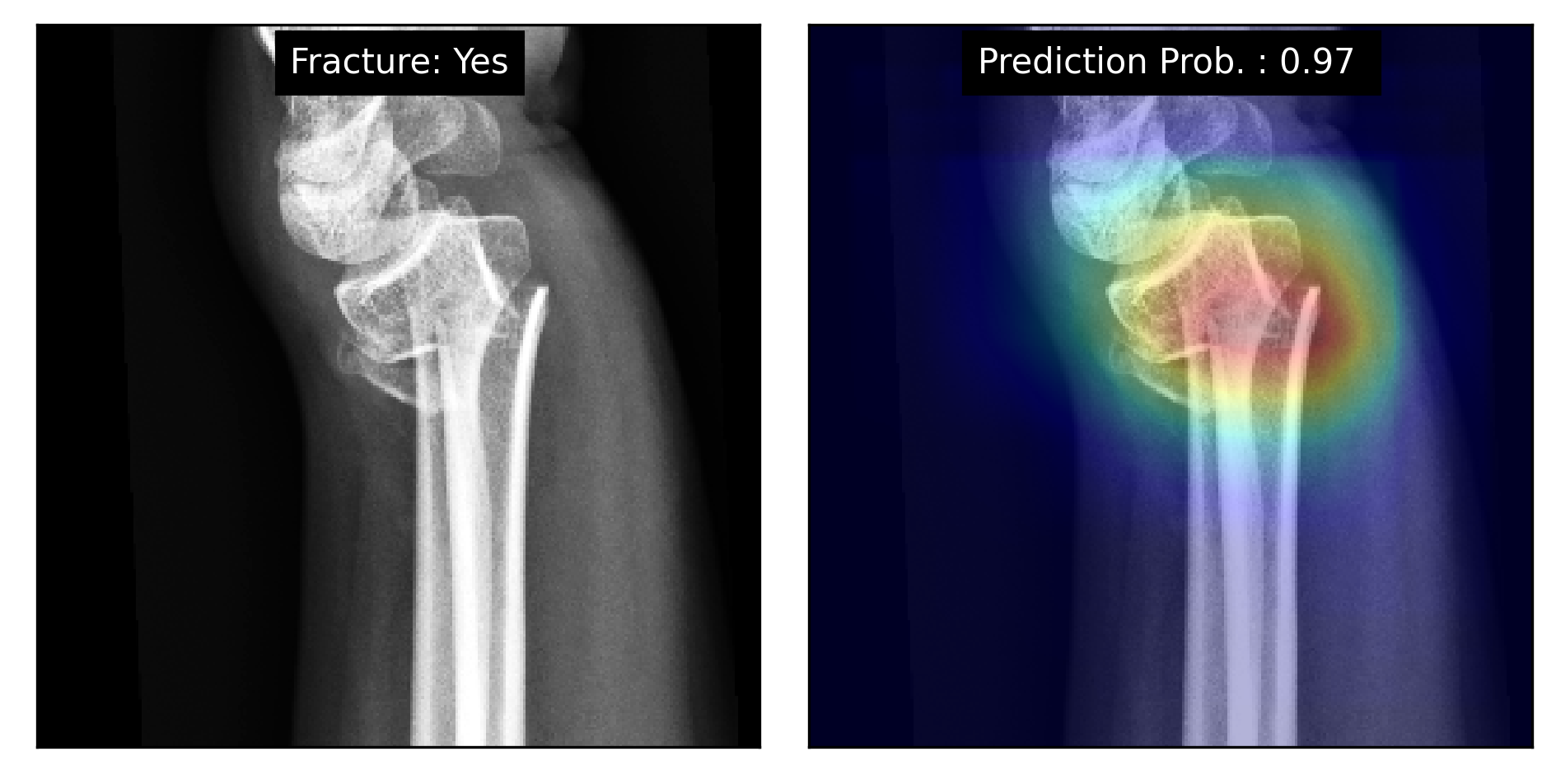}}

\subfloat[ \label{fig:gc_pa_hard}]{\includegraphics[width=.45\linewidth]{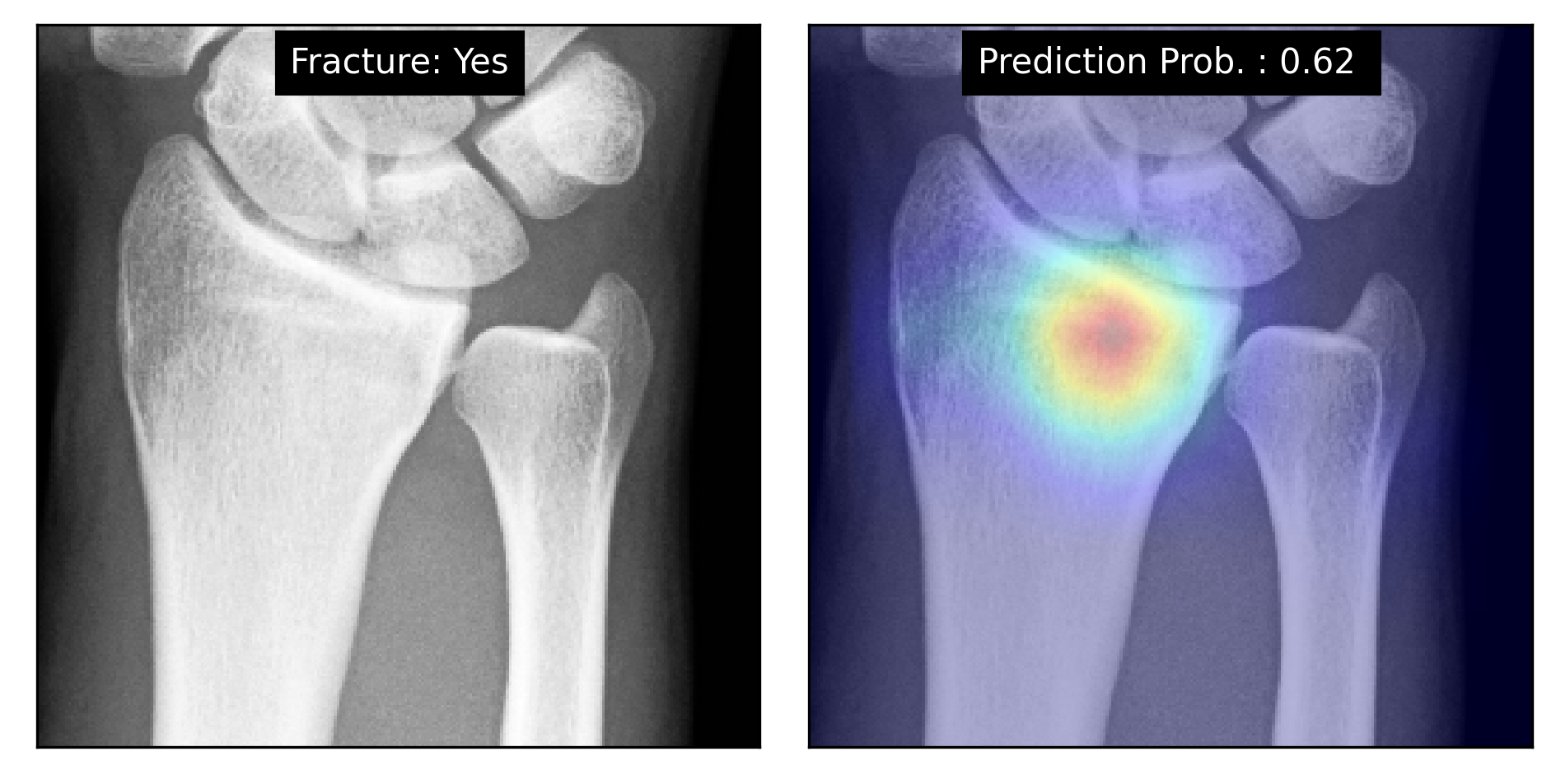}}
\hfil
\subfloat[ \label{fig:gc_lat_hard}]{\includegraphics[width=.45\linewidth]{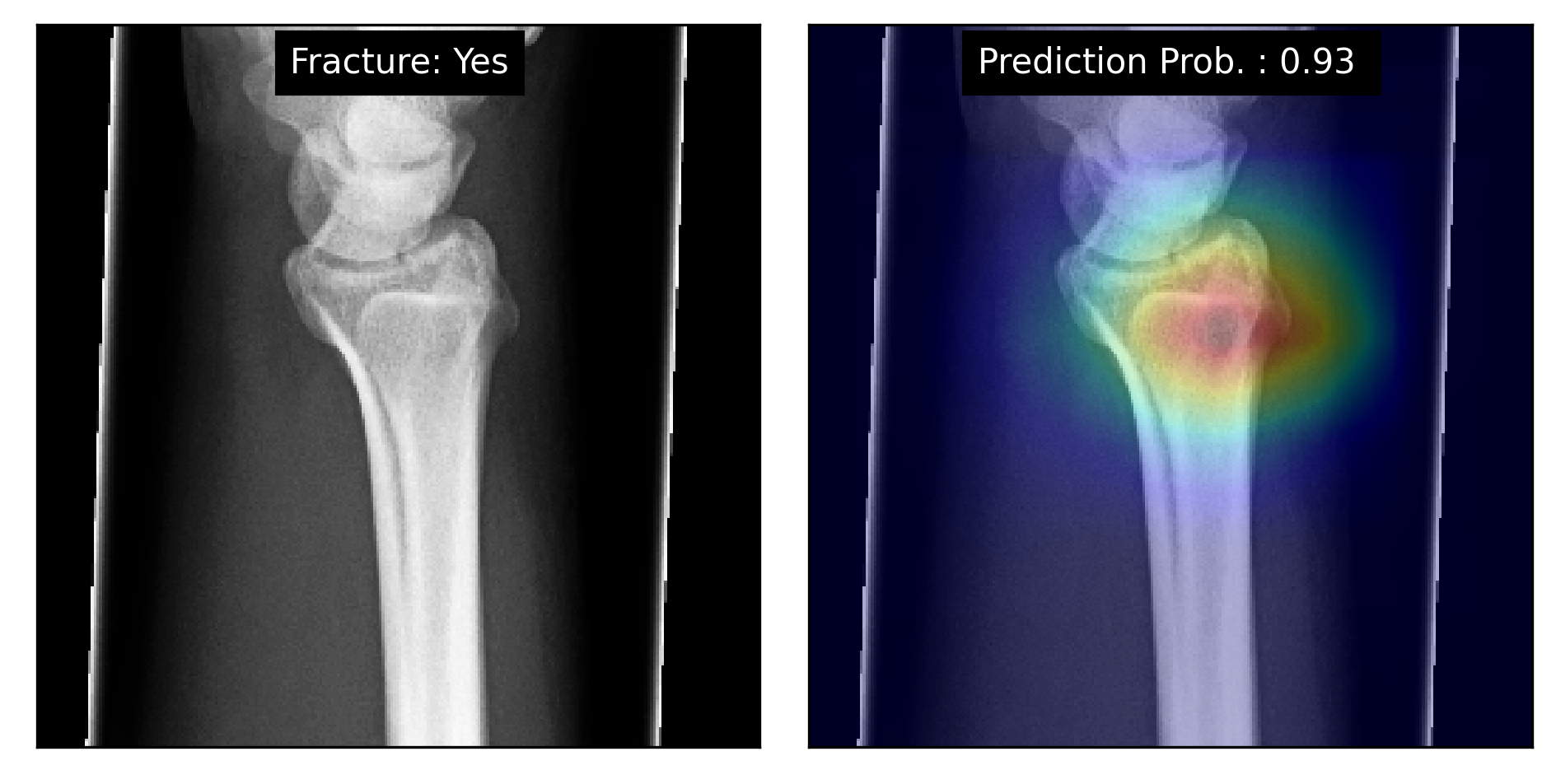}}

\subfloat[    \label{fig:fn_pa_hard}]{\includegraphics[width=.45\linewidth]{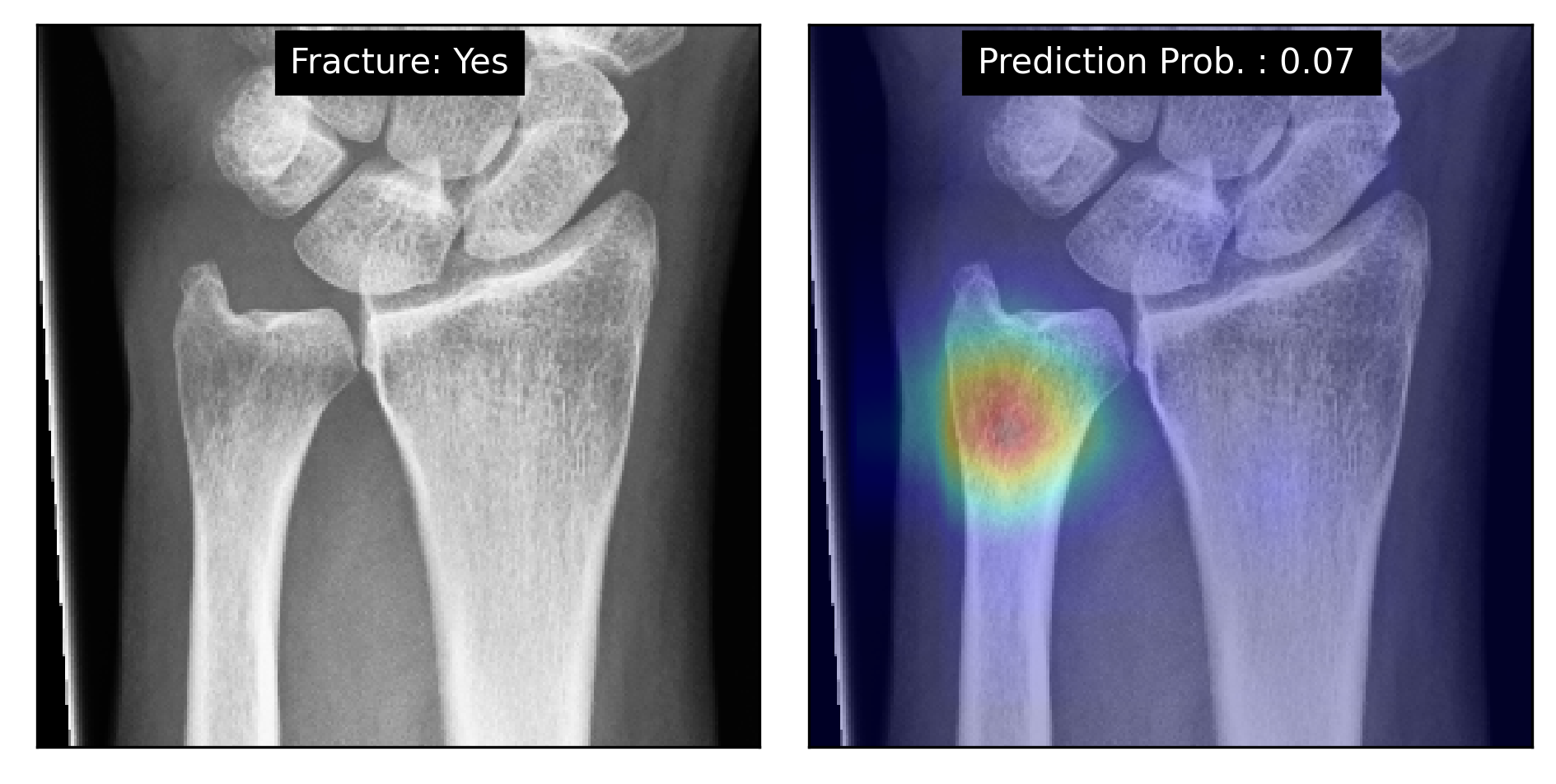}}
\hfil
\subfloat[   \label{fig:fn_lat_hard}]{\includegraphics[width=.45\linewidth]{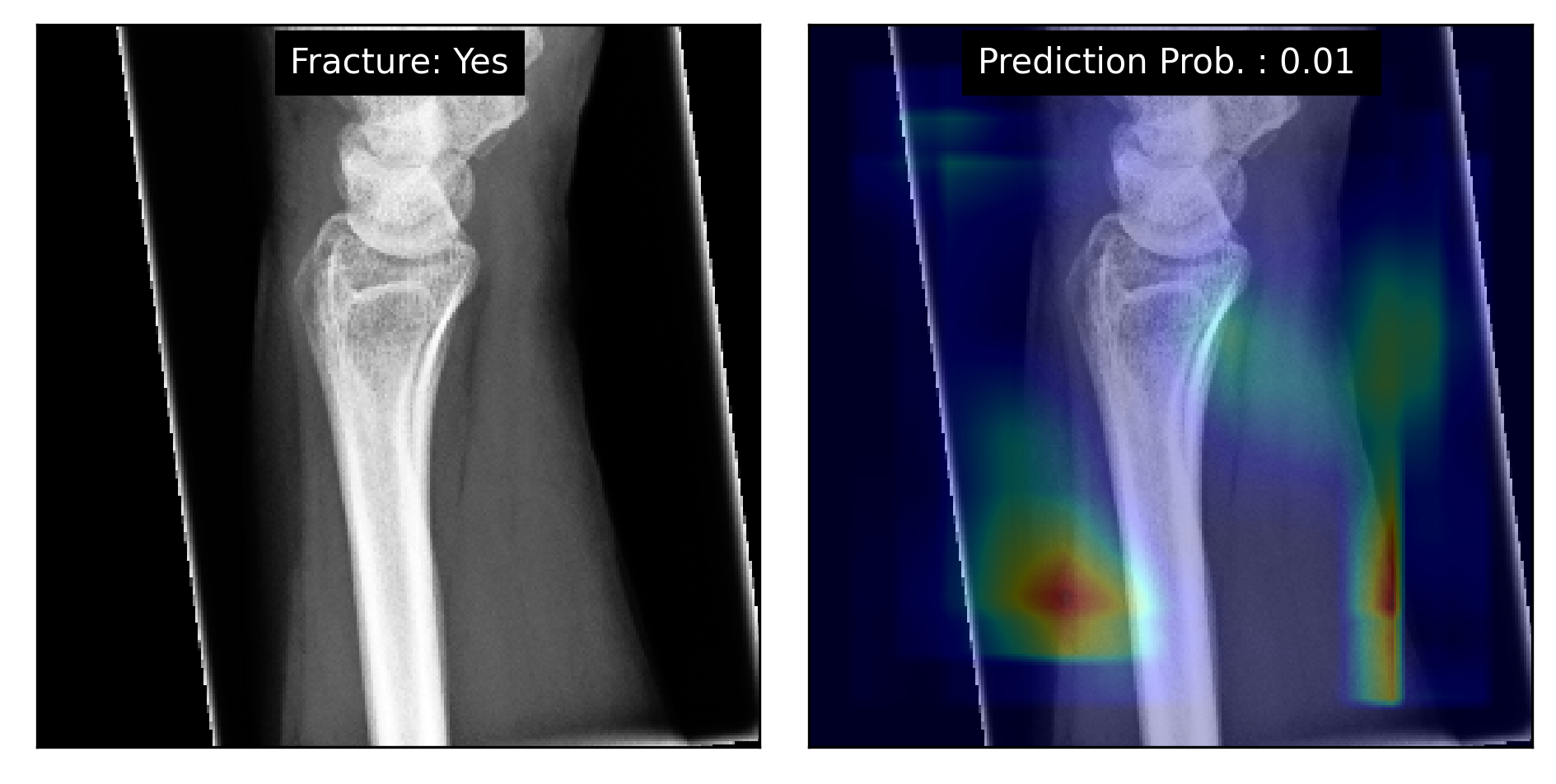}}

\caption{GradCAM-based heatmaps for the developed model. Each sub-figure shows input image and its label on the left side and GradCAM and prediction probability on the right side. (a) and (b) show PA and LAT views of a True Positive case from test set \#1 where the fracture is easily visible. (c) and (d) show both views of a True Positive from test set \#2 where the fracture is hardly visible and (e) and (f) show both views of a False Negative case from test set \#2 where the pipeline predicts them as normal.}
\label{fig:gradcam}
\end{figure}

In addition to the statistical analyses, we visualized the GradCAM-based heatmaps (\autoref{fig:gradcam}). For the True Positive cases in both datasets, on subplots (a)-(d), DeepWrist identified the correct zones, where distal radius fractures appear. The subplots (e) and (f) show that the model could not see these fractures, as they were not visually present in the image. 

\paragraph{Is there a distribution shift between general and hard cases?} 
Our results, show that for a 9 model Deep Ensemble, AUROC for OOD detection using predictive variance as uncertainty is $0.62\, (0.56 - 0.68)$, which indicates that the hard cases are not well detected as OOD with reliable performance. For ensembles with a lower number of models, we observed a similar or worse performance. Further insights are shown in Supplementary Section S4. 

%% file: tables/performance.tex
\begin{table}[ht]
\small
\centering
\begin{tabular}{ccccccccc}
\toprule
\bfseries{Dataset} & \bfseries{\multilinecell{Model}} & \bfseries{\multilinecell{AUROC \\ (95\% CI)}} & \bfseries{\multilinecell{AUPR \\ (95\% CI)}} & \bfseries{\multilinecell{Sensitivity,  \\ Recall, \\ TPR \\ (95\% CI)}} & \bfseries{\multilinecell{Specificity, \\ Selectivity, \\ TNR \\ (95\% CI)}} & \bfseries{\multilinecell{Precision \\ PPV \\ (95\% CI)}} & \bfseries{\multilinecell{$F_1$ Score \\ (95\% CI)}} & \bfseries{\multilinecell{BA \\ (95\% CI)}}\\ \midrule
\multirow{6}{*}{Test set \#1} & PA & \multilinecell{0.98 \\ (0.97 - 0.99)} & \multilinecell{0.99 \\ (0.98 - 0.99)} & \multilinecell{0.97 \\ (0.94 - 1.00)} & \multilinecell{0.88 \\ (0.80 - 0.94)} & \multilinecell{0.93 \\ (0.89 - 0.96)} & \multilinecell{0.95 \\ (0.92 - 0.97)} & \multilinecell{0.93 \\ (0.89 - 0.96)}\\ \cmidrule{2-9}
 & LAT & \multilinecell{0.98 \\ (0.97 - 0.99)} & \multilinecell{0.99 \\ (0.98 - 0.99)} & \multilinecell{0.97 \\ (0.94 - 1.00)} & \multilinecell{0.91 \\ (0.84 - 0.96)} & \multilinecell{0.94 \\ (0.91 - 0.97)} & \multilinecell{0.96 \\ (0.93 - 0.98)} & \multilinecell{0.94 \\ (0.90 - 0.97)}\\ \cmidrule{2-9}
 & Ensemble & \multilinecell{0.99 \\ (0.98 - 0.99)} & \multilinecell{0.99 \\ (0.99 - 0.99)} & \multilinecell{0.97 \\ (0.94 - 1.00)} & \multilinecell{0.87 \\ (0.79 - 0.93)} & \multilinecell{0.92 \\ (0.88 - 0.96)} & \multilinecell{0.95 \\ (0.92 - 0.97)} & \multilinecell{0.92 \\ (0.88 - 0.96)}\\ \midrule
\multirow{6}{*}{Test set \#2} & PA & \multilinecell{0.81 \\ (0.69 - 0.91)} & \multilinecell{0.61 \\ (0.44 - 0.80)} & \multilinecell{0.50 \\ (0.30 - 0.70)} & \multilinecell{0.89 \\ (0.82 - 0.95)} & \multilinecell{0.52 \\ (0.33 - 0.73)} & \multilinecell{0.51 \\ (0.31 - 0.68)} & \multilinecell{0.69 \\ (0.58 - 0.80)}\\ \cmidrule{2-9}
 & LAT & \multilinecell{0.83 \\ (0.70 - 0.93)} & \multilinecell{0.57 \\ (0.41 - 0.80)} & \multilinecell{0.50 \\ (0.30 - 0.70)} & \multilinecell{0.94 \\ (0.88 - 0.98)} & \multilinecell{0.66 \\ (0.46 - 0.90)} & \multilinecell{0.57 \\ (0.36 - 0.75)} & \multilinecell{0.72 \\ (0.60 - 0.83)}\\ \cmidrule{2-9}
 & Ensemble & \multilinecell{0.84 \\ (0.72 - 0.93)} & \multilinecell{0.64 \\ (0.46 - 0.83)} & \multilinecell{0.60 \\ (0.40 - 0.80)} & \multilinecell{0.92 \\ (0.87 - 0.97)} & \multilinecell{0.66 \\ (0.48 - 0.87)} & \multilinecell{0.63 \\ (0.44 - 0.80)} & \multilinecell{0.76 \\ (0.65 - 0.87)}\\ \bottomrule
\end{tabular}
\caption{DeepWrist's performance on trivial cases (test set \#1) and hard cases ( test set \#2). Here, AUROC is Area Under the receiver operating characteristic, AUPR is the Area Under Precision Recall curve,   CI is Confidence Interval, 95\% CI is shown in parentheses,  TPR is True Positive Rate, TNR is True Negative Rate and PPV is Positive Predictive Value and BA stands for Balanced Accuracy.}
\label{tab:test_perf}
\end{table}

%% file: tables/easy_set_stat.tex
\begin{table}[ht!]
\small
\centering
\begin{tabular}{m{0.107\textwidth}m{0.107\textwidth}m{0.107\textwidth}m{0.107\textwidth}m{0.107\textwidth}m{0.107\textwidth}m{0.107\textwidth}}
\toprule
\bfseries{} & \bfseries{\multilinecell{Radiology \\ Resident}} & \bfseries{Radiologist 1} & \bfseries{Radiologist 2} & \bfseries{\multilinecell{Primary Care \\ Physician 1}} & \bfseries{\multilinecell{Primary Care \\ Physician 2}} & \bfseries{\multilinecell{DeepWrist }}\\ \midrule
\multilinecell{Sensitivity \\ (95\% CI)} & \multilinecell{0.98 \\ (0.96 - 1.00)} & \multilinecell{1.00 \\ (1.00 - 1.00)} & \multilinecell{0.99 \\ (0.97 - 1.00)} & \multilinecell{0.99 \\ (0.97 - 1.00)} & \multilinecell{0.92 \\ (0.87 - 0.96)} & \multilinecell{0.97 \\ (0.94 - 1.00)}\\ \midrule
\multilinecell{Specificity \\ (95\% CI)} & \multilinecell{0.93 \\ (0.87 - 0.98)} & \multilinecell{0.97 \\ (0.93 - 1.00)} & \multilinecell{1.00 \\ (1.00 - 1.00)} & \multilinecell{0.73 \\ (0.62 - 0.82)} & \multilinecell{0.97 \\ (0.93 - 1.00)} & \multilinecell{0.87 \\ (0.79 - 0.93)}\\ \midrule
\multilinecell{Precision \\ (95\% CI)} & \multilinecell{0.96 \\ (0.92 - 0.99)} & \multilinecell{0.98 \\ (0.96 - 1.00)} & \multilinecell{1.00 \\ (1.00 - 1.00)} & \multilinecell{0.85 \\ (0.81 - 0.90)} & \multilinecell{0.98 \\ (0.95 - 1.00)} & \multilinecell{0.92 \\ (0.88 - 0.96)}\\ \midrule
\multilinecell{$F_1$ Score \\ (95\% CI)} & \multilinecell{0.97 \\ (0.95 - 0.99)} & \multilinecell{0.99 \\ (0.98 - 1.00)} & \multilinecell{0.99 \\ (0.98 - 1.00)} & \multilinecell{0.92 \\ (0.89 - 0.94)} & \multilinecell{0.95 \\ (0.92 - 0.97)} & \multilinecell{0.95 \\ (0.92 - 0.97)}\\ \midrule
\multilinecell{BA \\ (95\% CI)} & \multilinecell{0.96 \\ (0.92 - 0.98)} & \multilinecell{0.98 \\ (0.96 - 1.00)} & \multilinecell{0.99 \\ (0.98 - 1.00)} & \multilinecell{0.86 \\ (0.81 - 0.91)} & \multilinecell{0.94 \\ (0.91 - 0.97)} & \multilinecell{0.92 \\ (0.88 - 0.96)}\\ \bottomrule
\end{tabular}
\caption{Performance of 5 readers and DeepWrist on trivial cases (test set \#1). 95\% confidence intervals (CI) are shown in parentheses. BA stands for balanced accuracy.}
\label{tab:acc}
\end{table}

%% file: tables/hard_set_acc.tex
\begin{table}[ht!]
\small
\centering
\begin{tabular}{m{0.125\linewidth}m{0.125\linewidth}m{0.125\linewidth}m{0.125\linewidth}m{0.125\linewidth}m{0.125\linewidth}}
\toprule
\bfseries{} & \bfseries{Radiologist 1} & \bfseries{Radiologist 2} & \bfseries{\multilinecell{Primary Care \\ Physician 1}} & \bfseries{\multilinecell{Primary Care \\ Physician 2}} & \bfseries{\multilinecell{DeepWrist }}\\ \midrule

\multilinecell{Sensitivity \\ (95\% CI)} & \multilinecell{0.40 \\ (0.20 - 0.60)} & \multilinecell{0.40 \\ (0.20 - 0.60)} & \multilinecell{0.50 \\ (0.30 - 0.70)} & \multilinecell{0.60 \\ (0.40 - 0.80)} & \multilinecell{0.60 \\ (0.40 - 0.80)}\\ \midrule
\multilinecell{Specificity \\ (95\% CI)} & \multilinecell{0.95 \\ (0.90 - 0.98)} & \multilinecell{0.96 \\ (0.91 - 1.00)} & \multilinecell{0.80 \\ (0.71 - 0.88)} & \multilinecell{0.64 \\ (0.54 - 0.74)} & \multilinecell{0.92 \\ (0.87 - 0.97)}\\ \midrule
\multilinecell{Precision \\ (95\% CI)} & \multilinecell{0.66 \\ (0.41 - 0.91)} & \multilinecell{0.72 \\ (0.50 - 1.00)} & \multilinecell{0.37 \\ (0.23 - 0.52)} & \multilinecell{0.28 \\ (0.19 - 0.38)} & \multilinecell{0.66 \\ (0.48 - 0.87)}\\ \midrule
\multilinecell{$F_1$ Score \\ (95\% CI)} & \multilinecell{0.50 \\ (0.27 - 0.70)} & \multilinecell{0.51 \\ (0.28 - 0.70)} & \multilinecell{0.42 \\ (0.25 - 0.58)} & \multilinecell{0.38 \\ (0.25 - 0.50)} & \multilinecell{0.63 \\ (0.44 - 0.80)}\\ \midrule
\multilinecell{BA \\ (95\% CI)} & \multilinecell{0.67 \\ (0.57 - 0.78)} & \multilinecell{0.68 \\ (0.57 - 0.79)} & \multilinecell{0.65 \\ (0.53 - 0.76)} & \multilinecell{0.62 \\ (0.50 - 0.73)} & \multilinecell{0.76 \\ (0.65 - 0.87)}\\ \bottomrule
\end{tabular}
\caption{Performance of 4 readers and DeepWrist on hard cases (test set \#2). 95\% confidence intervals (CI) are shown in parentheses. BA stands for balanced accuracy.}
\label{tab:acc_ts2}
\end{table}

%% file: section/discussion.tex
\section*{Discussion}
In this study, we followed the recent works and trained a CNN-based pipeline for distal radius wrist fracture detection. Compared to recent studies on wrist fracture detection, on the general population dataset, our pipeline scored a better AUROC than others~\cite{bluthgen2020detection, thian2019convolutional, lindsey2018deep, kim2018artificial}. The important aim of our study, was to bring up the general issues of safety and robustness of AI in medical imaging to the attention of the reader. Earlier, this issue has been highlighted by Oakden-Rayner et \textit{al}.~\cite{oakden2020hidden}, and some results were shown on musculoskeletal image data from some of the public datasets. Our work is different from the prior art, as we investigated the problem on a real clinical dataset.

A novelty of our work is that we used the validation on challenging cases to expose the safety and robustness issues. In the medical AI domain, most of the studies (for example the fracture detection studies~\cite{bluthgen2020detection, thian2019convolutional, lindsey2018deep, kim2018artificial}) do not investigate the challenging cases in the evaluation. However, in a real clinical scenario, all kinds of cases (trivial, hard or with incidental findings) can appear. We showed that even in a relatively well studied domain, there exist issues of AI robustness, which expose the requirement for an additional algorithm safety assessment in the medical AI realm. 

On the general population test set (test set \#1), we observed a near perfect classification performance (AUPR: $0.99$, AUROC: $0.99$), which, however, still could not surpass the best human rater in terms of Sensitivity, Specificity, Precision, $F_1$ Score or Balanced Accuracy. The second set of experiments has shown a sharp downfall of performance for test set \#2. This dataset comprised the uncertain clinical cases, which could not be diagnosed by a radiologist from an X-ray image, and required an additional confirmation via CT imaging. We note that if we merge the uncertain cases with the general cases, the average performance remains still good, producing an AUROC of $0.97\, (0.95 - 0.98)$ and an AUPR of $0.97\,(0.96 - 0.98)$, matching the previous studies.

Along with the reported performance metrics, the inter-rater agreement analysis also shows similar results: all the raters have good agreement with the ground truth for the test set \#1, while disagreeing with the ground truth for test set \#2. In terms of fracture detection, Sensitivity, Precision, $F_1$ Score and Balanced Accuracy also decreased for all the raters on the test set \#2, indicating that it is difficult for humans to make the decision of the challenging cases.

We investigated deeper whether our model learned any significant associations, which are predictive of fractures on the test set \#2. We found that the predictions produced by our model are not more significant than the demographic variables on this dataset. This provides an opportunity for future studies to disentangle the prediction of fractures and the demographic variables.

Another aspect of our work is the assessment of the attention maps. We note that the GradCAM visualizations also confirmed that the DeepWrist did not find the signs of fractures in some of the images, and predicted the cases as negative, while the CT imaging diagnosed fracture. However, it was interesting to observe that the attention maps did not point at the locations of possible fractures. We believe the assessment of such attention maps in the future can tell about the prediction uncertainty, and could, perhaps, allow to detect the cases, which are likely to be misdiagnosed. When making automatic decisions in clinical practice, such information could be useful, as it could allow for automatic referral of the image to a radiologist, when a machine is incapable of making a decision. We note that similar ideas have been investigated in other domains, such as fundus imaging~\cite{leibig2017leveraging}, and we think that it is worth investigating them in the domain of musculoskeletal radiology. Our results show the attempt of using Deep Ensembles to quantify the total predictive uncertainty, however, we observed that the distinction between test set \#1 and test set \#2 is rather poor. We think different methods, which put a special focus on out-of-domain uncertainty may work better to analyze this problem.

Several limitations of this study should be mentioned. First, our training cases and the test set \#1 were annotated from the radiology reports, which might contain misdiagnosis. However, we tried to combat this limitation, by manually verifying the quality of the report during the annotation. In relation to this limitation, we note that the ground truth for the test set \#1 was derived from the consensus of R1 and R2, thereby yielding rather optimistic results in terms of the sensitivity and specificity. We think that future studies should also involve an independent set of readers, who will produce the ground truth. The second shortcoming of this work is that we had to exclude some of the cases from the statistical analysis due to their DICOM images having no age and sex metadata (see Supplementary Table S2). Therefore, we conducted the analysis of confounding factors using only the available data. The third limitation here is that the landmark annotations for training and the intra-rater variability analysis were done by a doctoral student (the first author). As a result, it is possible to have bias in the landmark annotation dataset. However, this limitation is rather minor, since after visual inspection of all our data processed by our landmark annotation method, we did not observe a single failure. The fourth limitation of the paper is that for the uncertainty estimation with Deep Ensembles, we were unable to use the power of transfer learning. Thereby, this could have affected the overall predictive performance of the ensemble. However we believe that despite this, the presented results are still indicative of how a state-of-the-art method for uncertainty estimation may perform in evaluating the domain shift. The fifth limitation of our work is limited data: the amount of challenging cases is much lower than the amount of general cases, and all data are taken from a single Hospital. We therefore think that the future studies need to conduct similar evaluations to ours across different hospitals and populations. The final, and major limitation of this work is that it rather poses a new challenge without proposing a solution for it. However, we considered the scope of this study to be in the realm of analysing the applicability of DL to the clinically challenging cases. As we already mentioned in the discussion of the attention maps, one could look at the uncertainty of predictions. The modern advances in Bayesian deep learning have potential to help with such matters~\cite{solovyev2020bayesian,farquhar2020radial}.

To conclude, we believe that the integration of AI into the clinical practice should be taken with care, and new requirements for regulatory approval may need to be introduced. We believe that our work opens a new avenue for research in the realm of DL, and we consider that new methods, which are capable of robust out-of-domain predictive uncertainty estimation are needed to ensure the safety of using AI in healthcare.

%% file: tables/data_stat.tex
\begin{longtable}{cccccccc}
\tiny
\centering
\endfirsthead
\toprule
\multicolumn{8}{c}{Continuation of Table from previous page}
\\ \hline
\bfseries{Dataset} & \bfseries{Label} & \bfseries{Gender} & \bfseries{Count} & \bfseries{Mean Age} & \bfseries{SD of Age} & \bfseries{Age Range} & \bfseries{\multilinecell{Number of  \\ Age Records}}\\ \midrule
\endhead
\hline
\endfoot
\endlastfoot
\toprule
\bfseries{Dataset} & \bfseries{Label} & \bfseries{Sex} & \bfseries{Count} & \bfseries{Mean Age} & \bfseries{SD of Age} & \bfseries{Age Range} & \bfseries{\multilinecell{Number of  \\ Age Records}}\\ \midrule
\multirow{6}{*}[-1.25em]{Training set} & \multirow{3}{*}[-0.5em]{Fracture} & Male & 252 & 48.23 & 18.51 & 15 - 89 & 206\\ \cmidrule{3-8}
 &  & Female & 696 & 60.51 & 17.04 & 15 - 94 & 585\\ \cmidrule{3-8}
 &  & Unknown & 5 & 47.50 & 13.94 & 27 - 66 & 4\\ \cmidrule{2-8}
 & \multirow{3}{*}[-0.5em]{Normal} & Male & 399 & 42.21 & 17.89 & 16 - 88 & 300\\ \cmidrule{3-8}
 &  & Female & 588 & 45.23 & 17.37 & 15 - 96 & 465\\ \cmidrule{3-8}
 &  & Unknown & 6 & 35.50 & 20.52 & 22 - 71 & 4\\ \midrule
\multirow{6}{*}[-1.25em]{Test set \#1} & \multirow{3}{*}[-0.5em]{Fracture} & Male & 22 & 50.45 & 21.43 & 18 - 84 & 22\\ \cmidrule{3-8}
 &  & Female & 105 & 64.21 & 16.58 & 22 - 93 & 104\\ \cmidrule{3-8}
 &  & Unknown & 2 & 62.50 & 7.50 & 55 - 70 & 2\\ \cmidrule{2-8}
 & \multirow{3}{*}[-0.5em]{Normal} & Male & 35 & 43.51 & 19.96 & 19 - 92 & 35\\ \cmidrule{3-8}
 &  & Female & 42 & 56.07 & 23.19 & 19 - 96 & 42\\ \cmidrule{3-8}
 &  & Unknown & 1 & 20.00 & 0.00 & 20 - 20 & 1\\ \midrule
\multirow{6}{*}[-1.25em]{Test set \#2} & \multirow{3}{*}[-0.5em]{Fracture} & Male & 13 & 48.75 & 15.71 & 23 - 72 & 8\\ \cmidrule{3-8}
 &  & Female & 7 & 53.80 & 17.42 & 20 - 70 & 5\\ \cmidrule{3-8}
 &  & Unknown & 0 & 0.00 & 0.00 & 0 - 0 & 0\\ \cmidrule{2-8}
 & \multirow{3}{*}[-0.5em]{Normal} & Male & 48 & 36.24 & 18.02 & 17 - 80 & 33\\ \cmidrule{3-8}
 &  & Female & 32 & 55.19 & 16.46 & 23 - 88 & 26\\ \cmidrule{3-8}
 &  & Unknown & 5 & 53.00 & 2.00 & 51 - 55 & 2\\ \bottomrule
\caption{Dataset Statistics. SD stands for Standard Deviation. The `Number of Age Records' column indicates the number of cases for which the age data is recorded.}
\label{tab:data_stat}
\end{longtable}